\documentclass[journal]{IEEEtran}

\usepackage{amsfonts}

\usepackage{adjustbox}
\usepackage{algorithmicx}
\usepackage{algpseudocode}
\usepackage[ruled]{algorithm}
\usepackage{graphicx}
\usepackage{caption}
\usepackage{cite}
\usepackage{graphicx,times,amsmath} 
\usepackage{caption}
\usepackage{subcaption}
\usepackage[rightcaption]{sidecap}
\usepackage{caption}
\usepackage{multirow}
\usepackage{hhline}
\usepackage{amsmath}
\usepackage{epstopdf}
\usepackage{tabularx}
\usepackage[font=footnotesize]{caption}

\usepackage{makecell}

\newcommand\norm[1]{\left\lVert#1\right\rVert}

\graphicspath{ {images/} }

\ifCLASSINFOpdf
\else
\fi

\begin{document}
\title{Recent Advances in Recurrent Neural Networks}

\author{Hojjat Salehinejad,
		Sharan Sankar,
        Joseph Barfett,
        Errol Colak,
         and Shahrokh Valaee
\thanks{H. Salehinejad is with the Department of Electrical \& Computer Engineering, University of Toronto, Toronto, Canada, and Department of Medical Imaging, St. Michael's Hospital, University of Toronto, Toronto, Canada, e-mail: salehinejadh@smh.ca.}
\thanks{S. Sankar is with the Department of Electrical and Computer Engineering, University of Waterloo, Waterloo, Canada, e-mail:sdsankar@edu.uwaterloo.ca.}
\thanks{J. Barfett and E. Colak are with the Department of Medical Imaging, St. Michael's Hospital, University of Toronto, Toronto, Canada, e-mail: \{barfettj,colake\}@smh.ca.}
\thanks{S. Valaee is with the Department of Electrical \& Computer Engineering, University of Toronto, Toronto, Canada, e-mail: valaee@ece.utoronto.ca.}
}

\markboth{}%
{Salehinejad \MakeLowercase{\textit{et al.}}: {Recent Advances in Recurrent Neural Networks}}
\maketitle

\begin{abstract}
Recurrent neural networks (RNNs) are capable of learning features and long term dependencies from sequential and time-series data. The RNNs have a stack of non-linear units where at least one connection between units forms a directed cycle. A well-trained RNN can model any dynamical system; however, training RNNs is mostly plagued by issues in learning long-term dependencies. In this paper, we present a survey on RNNs and several new advances for newcomers and professionals in the field. The fundamentals and recent advances are explained and the research challenges are introduced. 
\end{abstract}


\section{Introduction}
\label{sec:introduction}


\IEEEPARstart{A}{rtificial} neural networks (ANNs) are made from layers of connected units called artificial neurons. A ``shallow network" refers to an ANN with one input layer, one output layer, and at most one hidden layer without a recurrent connection. As the number of layers increases, the complexity of network increases too. More number of layers or recurrent connections generally increases the depth of the network and empowers it to provide various levels of data representation and feature extraction, referred to as ``deep learning". In general, these networks are made from nonlinear but simple units, where the higher layers provide a more abstract representation of data and suppresses unwanted variability \cite{lecun2015deep}. Due to optimization difficulties caused by composition of the nonlinearity at each layer, not much work occurred on deep network architectures before significant advances in 2006~\cite{hinton2006fast},~\cite{bengio2013advances}.
\begin{table}
\vspace{6pt}
\caption{Some of the major advances in recurrent neural networks (RNNs) at a glance.}
\begin{center}
\footnotesize
\begin{tabular}{|c|c|l|}

\hline
Year&First Author&\multicolumn{1}{|c|}{Contribution}\\
\hhline{---}
 \hline
1990	&  Elman      & Popularized simple RNNs (Elman network)\\ \hline
1993	&  Doya	    &Teacher forcing for gradient descent (GD)\\ \hline
1994&  Bengio	    &\makecell[l]{Difficulty in learning long term dependencies \\with gradient descend }\\ \hline
1997	&Hochreiter   &\makecell[l]{LSTM: long-short term memory\\ for vanishing gradients problem}\\ \hline
1997 & Schuster  & BRNN: Bidirectional recurrent neural networks\\ \hline
1998	&  LeCun	     &\makecell[l]{Hessian matrix approach for\\ vanishing gradients problem}\\ \hline	
2000 & Gers    & Extended LSTM with forget gates	\\ \hline	
2001	&  Goodman &Classes for fast Maximum entropy training\\ \hline		
2005	&  Morin	      &\makecell[l]{A hierarchical softmax function for\\ language modeling using RNNs}\\ \hline
2005 & Graves   & BLSTM: Bidirectional LSTM\\ \hline
2007&	Jaeger	&Leaky integration neurons\\ \hline
2007 & Graves & MDRNN: Multi-dimensional RNNs\\ \hline		
2009	&  Graves	     &LSTM for hand-writing recognition \\ \hline
2010	&  Mikolov      &	RNN based language model\\ \hline
2010	&  Neir     &	\makecell[l]{Rectified linear unit (ReLU) for\\ vanishing gradient problem}\\ \hline
2011	&  Martens      &Learning RNN with Hessian-free optimization\\ \hline
2011	&  Mikolov     &	\makecell[l]{RNN by back-propagation through\\ time (BPTT)  for statistical language modeling}\\ \hline
2011	&  Sutskever  &	\makecell[l]{Hessian-free optimization with\\ structural damping}\\ \hline
2011	&  Duchi         &	Adaptive learning rates for each weight\\ \hline
2012	&  Gutmann	&Noise-contrastive estimation (NCE)\\ \hline
2012	&  Mnih	      &\makecell[l]{NCE for training neural probabilistic\\
language models (NPLMs)}\\ \hline
2012	&  Pascanu	&\makecell[l]{Avoiding exploding gradient problem\\ by gradient clipping}\\ \hline
2013&  Mikolov	       &\makecell[l]{Negative sampling instead of\\ hierarchical softmax}\\ \hline
2013	&  Sutskever   &	\makecell[l]{Stochastic gradient descent (SGD) \\with momentum}\\ \hline    
2013 & Graves & Deep LSTM RNNs (Stacked LSTM) \\ \hline
2014 & Cho & Gated recurrent units \\ \hline
2015 &  Zaremba    & Dropout for reducing Overfitting   \\ \hline       
2015	&  Mikolov      &	\makecell[l]{Structurally constrained recurrent network\\ (SCRN) to enhance learning longer memory\\ for vanishing gradient problem} \\ \hline
2015 & Visin & \makecell[l]{ReNet: A RNN-based alternative to \\convolutional neural networks}       \\ \hline  
2015 & Gregor & DRAW: Deep recurrent attentive writer \\ \hline     
2015 & Kalchbrenner & Grid long-short term memory\\ \hline      
2015 & Srivastava  & Highway network \\ \hline    
2017 & Jing & Gated orthogonal recurrent units  \\ \hline 
\end{tabular}
\label{T:RNN_history}
\end{center}
\vspace{-9pt}
\end{table}
ANNs with recurrent connections are called recurrent neural networks (RNNs), which are capable of modelling sequential data for sequence recognition and prediction \cite{bengio1994learning}. RNNs are made of high dimensional hidden states with non-linear dynamics \cite{sutskever2011generating}. The structure of hidden states work as the memory of the network and state of the hidden layer at a time is conditioned on its previous state \cite{mikolov2014learning}. This structure enables the RNNs to store, remember, and process past complex signals for long time periods. RNNs can map an input sequence to the output sequence at the current timestep and predict the sequence in the next timestep. 

A large number of papers are published in the literature based on RNNs, from architecture design to applications development. 
In this paper, we focus on discussing discrete-time RNNs and recent advances in the field. Some of the major advances in RNNs through time are listed in Table~\ref{T:RNN_history}. The development of back-propagation using gradient descent (GD) has provided a great opportunity for training RNNs. This simple training approach has accelerated practical achievements in developing RNNs \cite{sutskever2011generating}. However, it comes with some challenges in modelling long-term dependencies such as vanishing and exploding gradient problems, which are discussed in this paper.

The rest of paper is organized as follow. The fundamentals of RNNs are presented in Section~\ref{sec:srnn}. Methods for training RNNs are discussed in Section~\ref{sec:trainingRNN} and a variety of RNNs architectures are presented in Section~\ref{sec:RNNArch}. 
The regularization methods for training RNNs are discussed in Section~\ref{sec:regularization}. Finally, a brief survey on major applications of RNN in signal processing is presented in Section~\ref{sec:RNNsigapps}.

\section{A Simple Recurrent Neural Network}
\label{sec:srnn}
RNNs are a class of supervised machine learning models, made of artificial neurons with one or more feedback loops~\cite{haykin1994neural}. The feedback loops are recurrent cycles over time or sequence (we call it time throughout this paper) \cite{mikolov2010recurrent}, as shown in Figure~\ref{fig:srnn}. Training a RNN in a supervised fashion requires a training dataset of input-target pairs. The objective is to minimize the difference between the output and target pairs (i.e., the loss value) by optimizing the weights of the network.

\subsection{Model Architecture}
\begin{figure}[!tp]
\footnotesize
\centering
        \begin{subfigure}[b]{0.28\textwidth}
           \centering 
                 \includegraphics[width=1\linewidth]{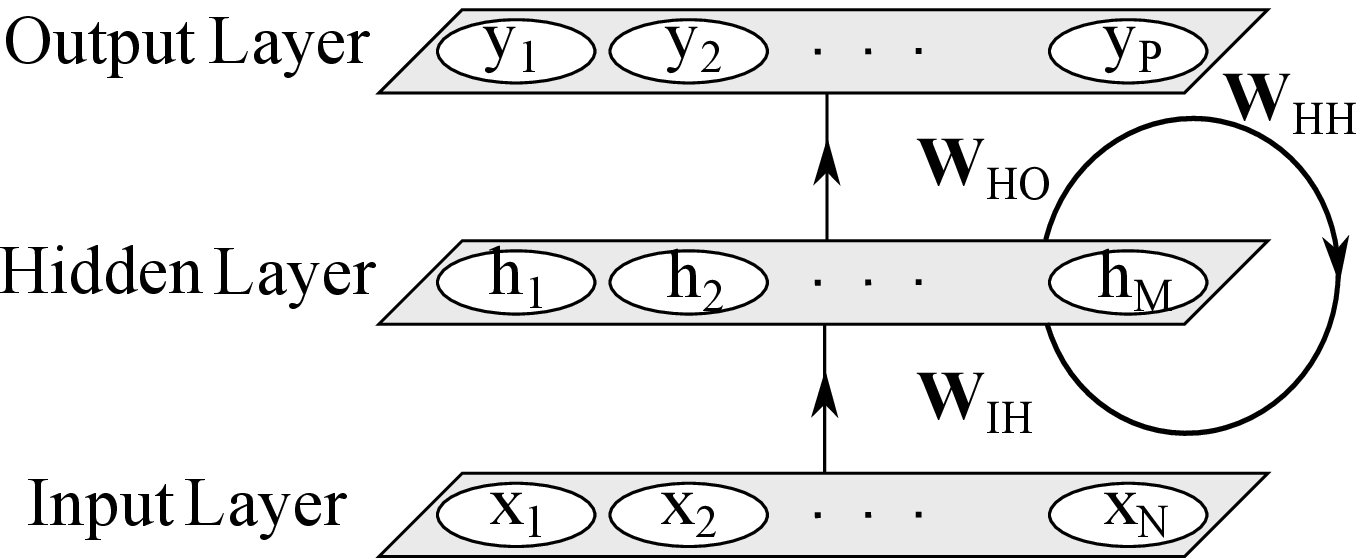}
                \caption{Folded RNN.}
                \label{fig:srnn_folded}
        \end{subfigure}%
  
  \vspace{0.2cm}
 \begin{subfigure}[b]{0.45\textwidth}
    \centering 
                 \includegraphics[width=1\linewidth]{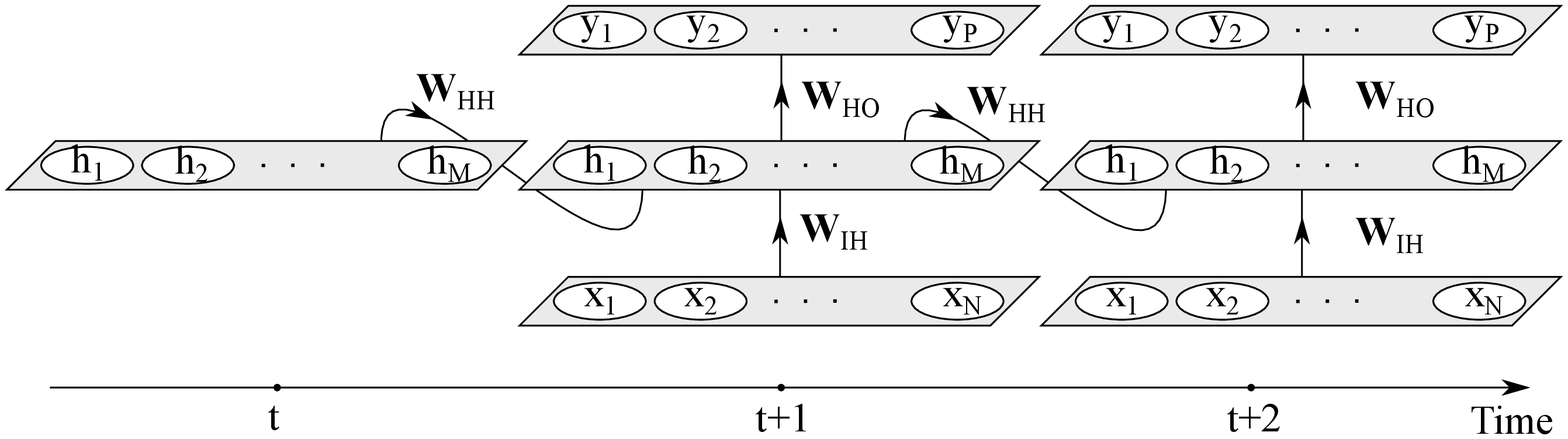}
                \caption{Unfolded RNN through time.}
                \label{fig:srnn_unfolded}
        \end{subfigure}%
~
        \caption{A simple recurrent neural network (RNN) and its unfolded structure through time $t$. Each arrow shows a full connection of units between the layers. To keep the figure simple, biases are not shown.}
                \label{fig:srnn}     
\end{figure}

A simple RNN has three layers which are input, recurrent hidden, and output layers, as presented in Figure~\ref{fig:srnn_folded}. The input layer has $N$ input units. The inputs to this layer is a sequence of vectors through time $t$ such as 
$\{..., \textbf{x}_{t-1}, \textbf{x}_{t}, \textbf{x}_{t+1},...\}$, where $\textbf{x}_{t}=(x_{1}, x_{2}, ..., x_{N})$. The input units in a fully connected RNN are connected to the hidden units in the hidden layer, where the connections are defined with a weight matrix $\textbf{W}_{IH}$. The hidden layer has $M$ hidden units $\textbf{h}_{t}=(h_{1}, h_{2}, ..., h_{M})$, that are connected to each other through time with recurrent connections, Figure~\ref{fig:srnn_unfolded}. The initialization of hidden units using small non-zero elements can improve overall performance and stability of the network \cite{sutskever2013importance}. The hidden layer defines the state space or ``memory'' of the system as

\begin{equation}
\textbf{h}_{t} = f_{H}(\textbf{o}_{t}),
\label{eq:SRNN_hidden_state}
\end{equation}
where 
\begin{equation}
\textbf{o}_{t}=\textbf{W}_{IH}\textbf{x}_{t}+\textbf{W}_{HH}\textbf{h}_{t-1}+\textbf{b}_{h},
\label{eq:SRNN_hidden_output}
\end{equation}
$f_{H}(\cdot)$ is the hidden layer activation function, and $\textbf{b}_{h}$ is the bias vector of the hidden units. The hidden units are connected to the output layer with weighted connections $\textbf{W}_{HO}$. The output layer has $P$ units $\textbf{y}_{t}=(y_{1}, y_{2}, ..., y_{P})$ that are computed as
\begin{equation}
\textbf{y}_{t} = f_{O}(\textbf{W}_{HO}\textbf{h}_{t}+\textbf{b}_{o})
\label{eq:SRNN_outcome}
\end{equation}
where $f_{O}(\cdot)$ is the activation functions and $\textbf{b}_{o}$ is the bias vector in the output layer. Since the input-target pairs are sequential through time, the above steps are repeated consequently over time $t=(1,...,T)$. The Eqs.~(\ref{eq:SRNN_hidden_state})~and~(\ref{eq:SRNN_outcome}) show a RNN is consisted of certain non-linear state equations, which are iterable through time. In each timestep, the hidden states provide a prediction at the output layer based on the input vector. The hidden state of a RNN is a set of values, which apart from the effect of any external factors, summarizes all the
unique necessary information about the past states of the network over many timesteps. This integrated information can define future behaviour of the network and make accurate predictions at the output layer \cite{sutskever2011generating}. A RNN uses a simple nonlinear activation function in every unit. However, such simple structure is capable of modelling rich dynamics, if it is well trained through timesteps. 

\subsection{Activation Function}
For linear networks, multiple linear hidden layers act as a single linear hidden layer \cite{bengio2007scaling}. Nonlinear functions are more powerful than linear functions as they can draw nonlinear boundaries. The nonlinearity in one or successive hidden layers in a RNN is the reason for learning input-target relationships. 

\begin{figure}
\footnotesize
\centering
        \begin{subfigure}[b]{0.25\textwidth}
           \centering 
                 \includegraphics[width=1\linewidth]{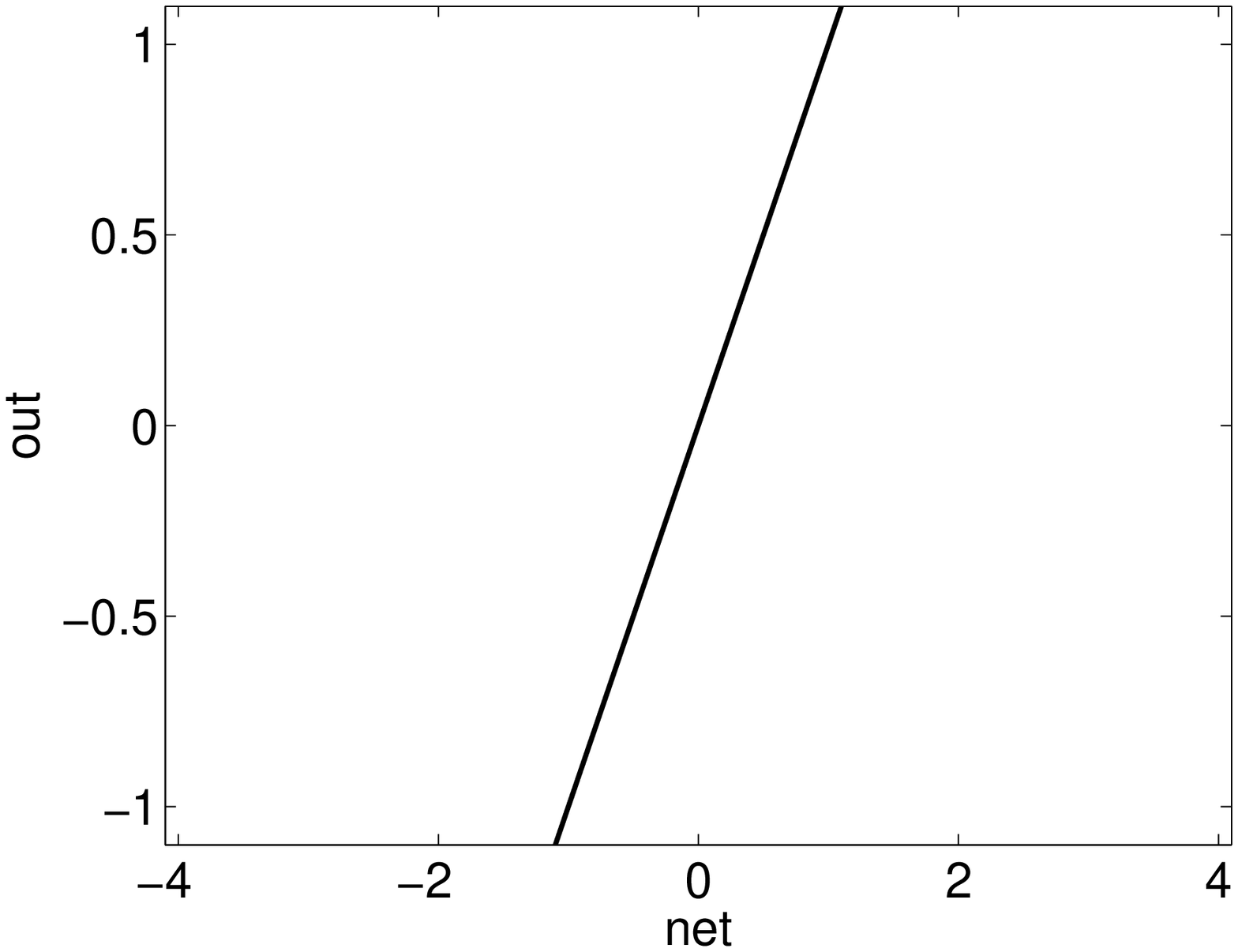}
                \caption{Linear.}
                \label{fig:deep_in_h1}
        \end{subfigure}%
 \begin{subfigure}[b]{0.25\textwidth}
    \centering 
                 \includegraphics[width=1\linewidth]{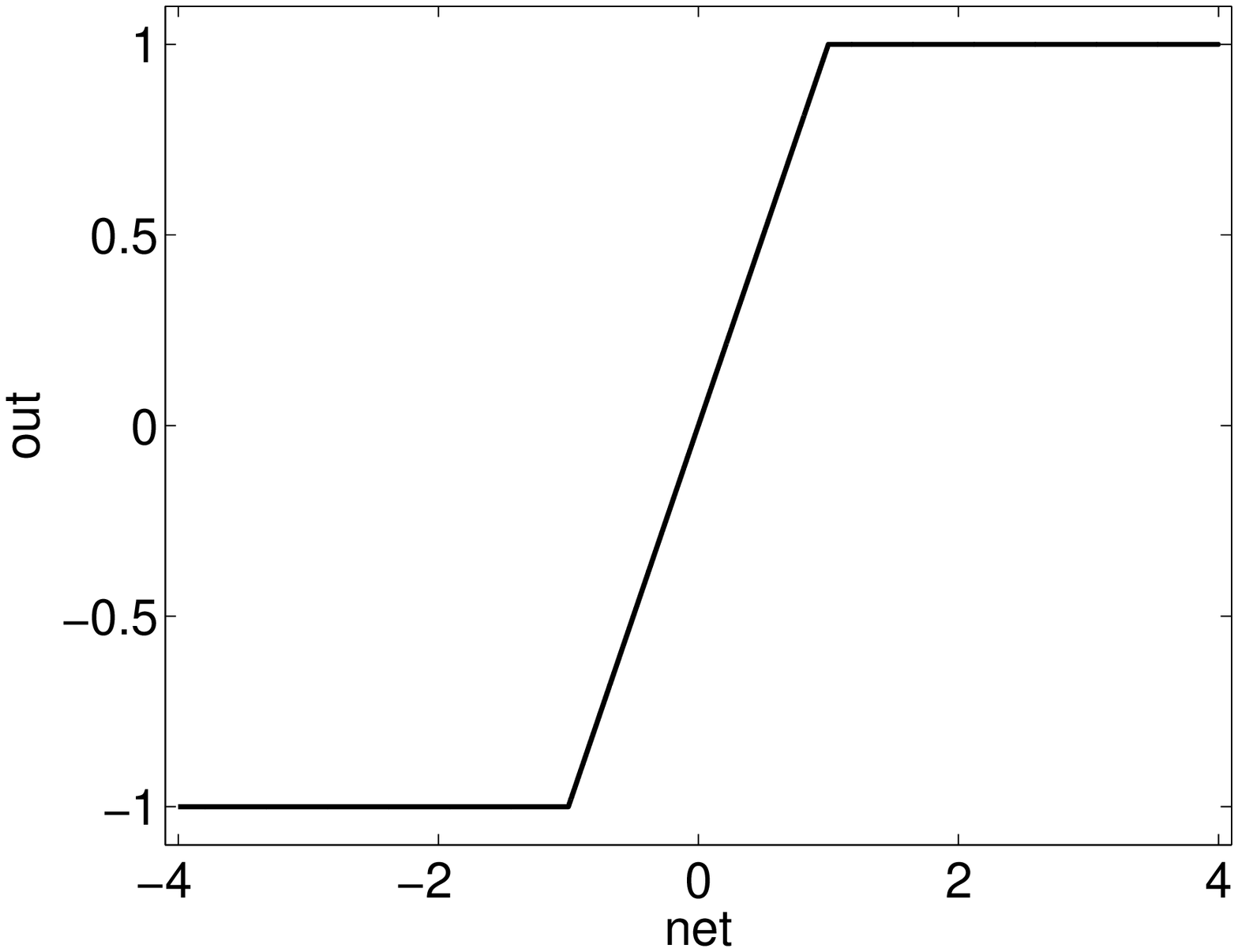}
                \caption{Piecewise linear.}
                \label{fig:deep_in_h2}
        \end{subfigure}%
        
 \begin{subfigure}[b]{0.25\textwidth}
    \centering 
                 \includegraphics[width=1\linewidth]{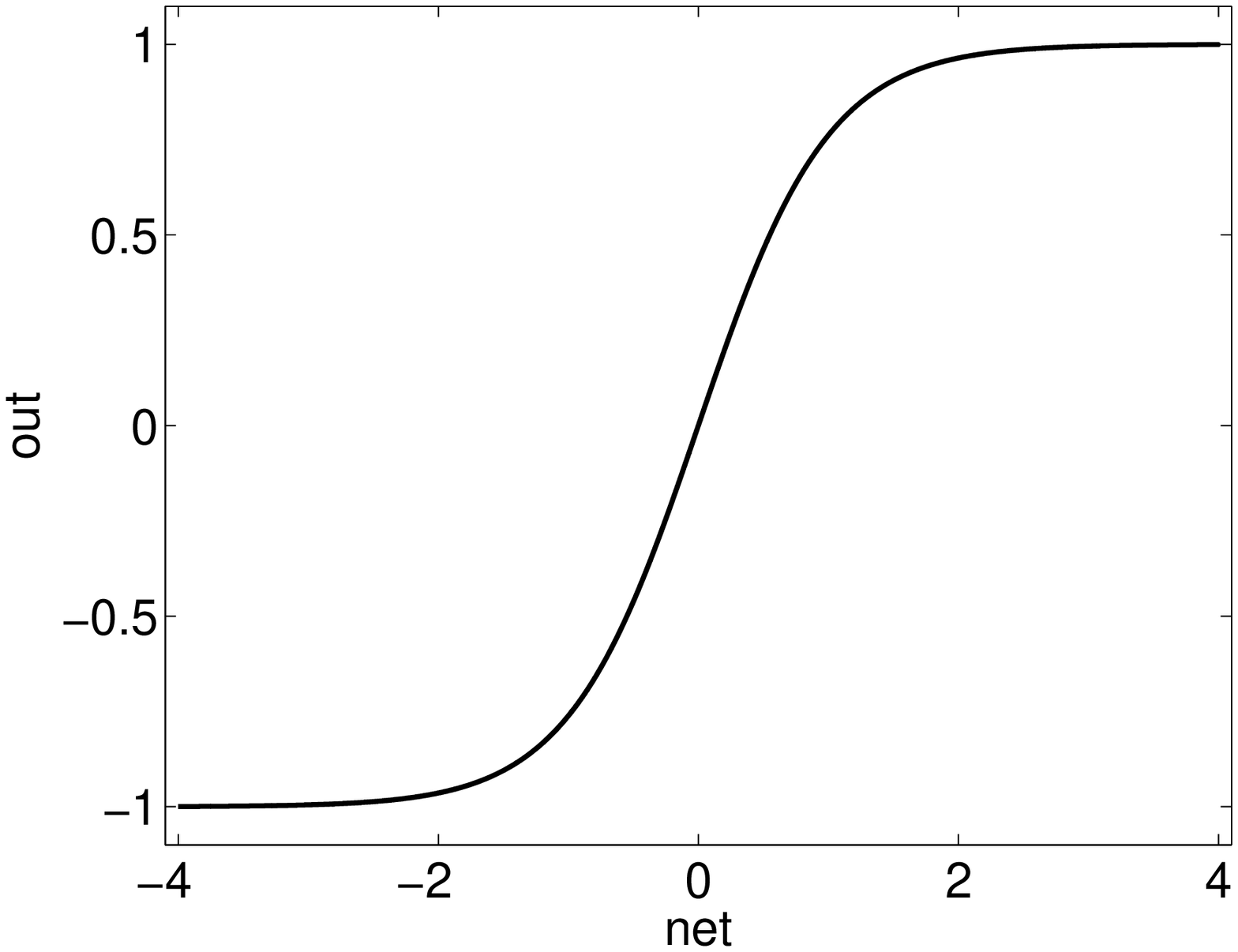}
                \caption{tanh(net).}
                \label{fig:deep_in_h3}
        \end{subfigure}%
\begin{subfigure}[b]{0.25\textwidth}
    \centering 
                 \includegraphics[width=1\linewidth]{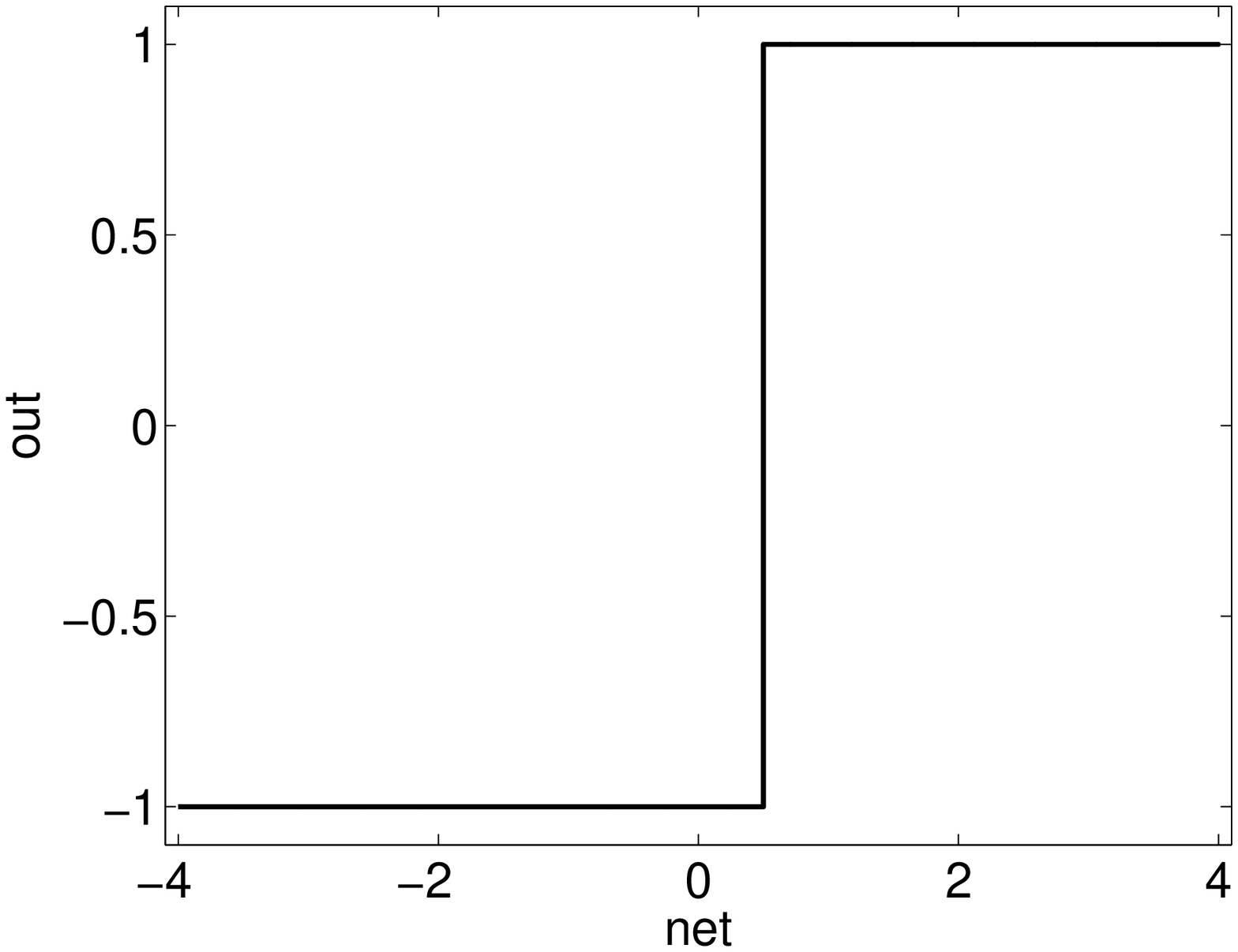}
                \caption{Threshold.}
                \label{fig:deep_in_h4}
        \end{subfigure}%
        
\begin{subfigure}[b]{0.25\textwidth}
    \centering 
                 \includegraphics[width=1\linewidth]{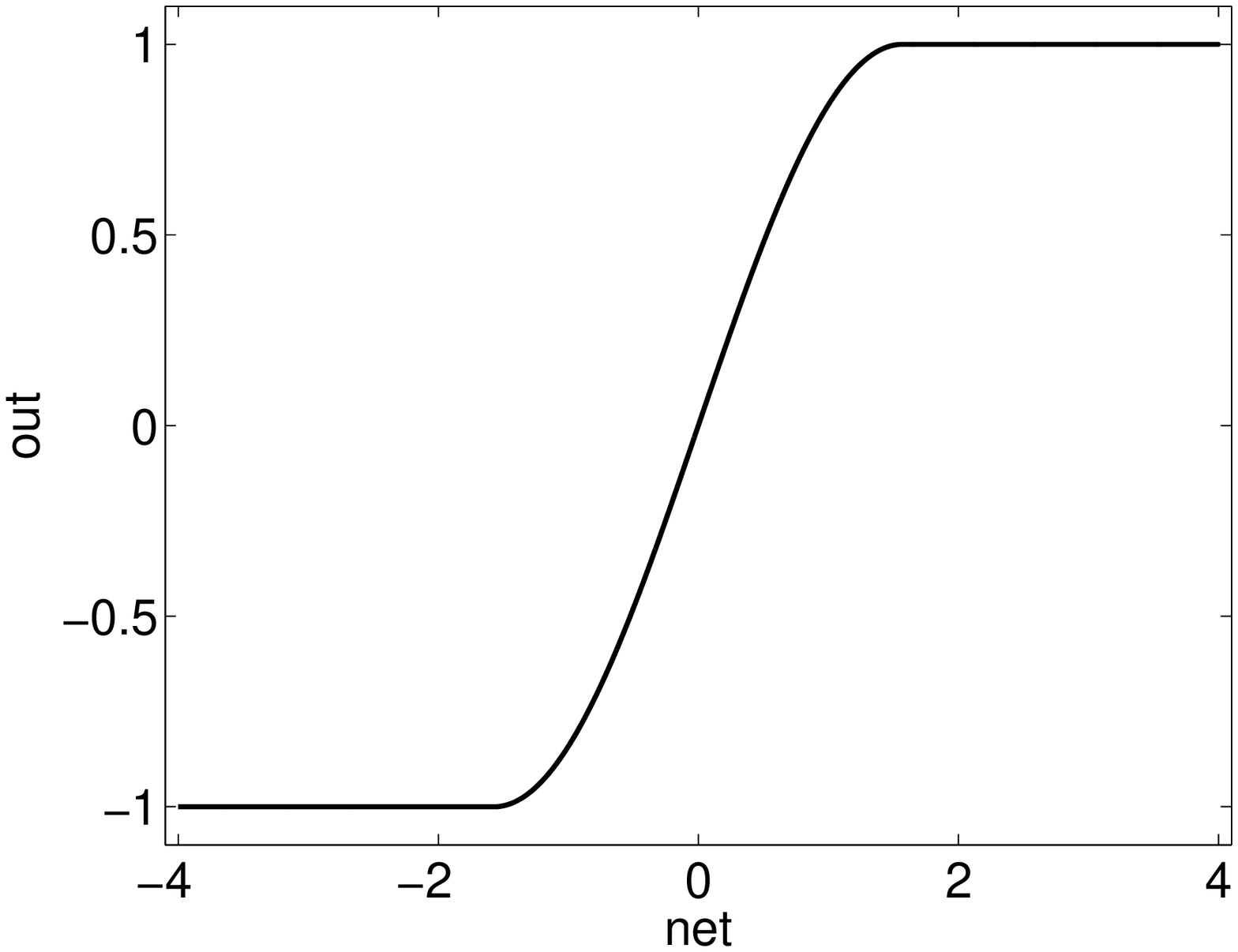}
                \caption{sin(net) until saturation.}
                \label{fig:deep_in_h4}
        \end{subfigure}%
        \begin{subfigure}[b]{0.25\textwidth}
    \centering 
                 \includegraphics[width=1\linewidth]{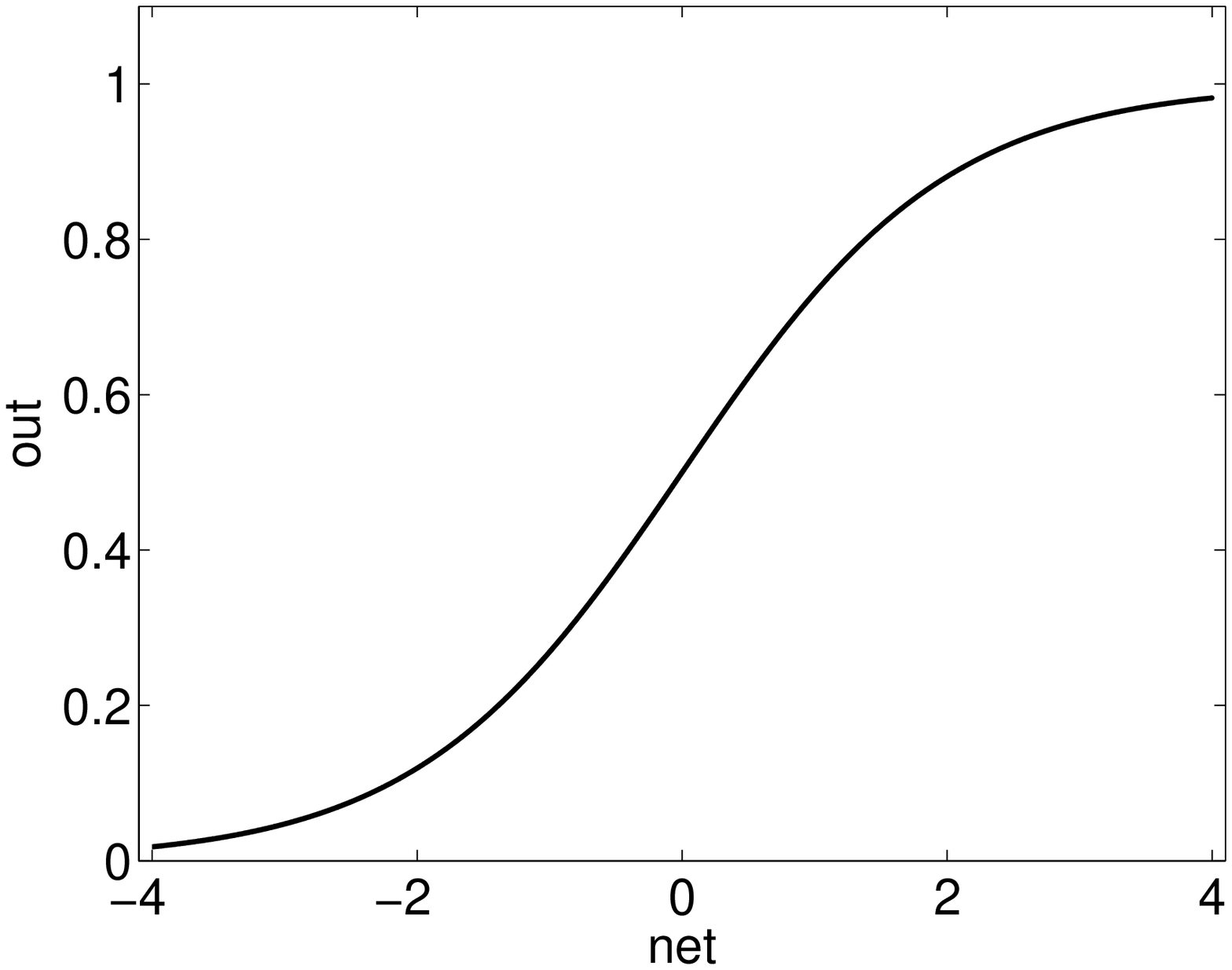}
                \caption{Sigmoid.}
                \label{fig:deep_in_h4}
        \end{subfigure}%

        \caption{Most common activation functions.}
                \label{fig:activation}     
\end{figure}

Some of the most popular activation functions are presented in Figure~\ref{fig:activation}. The ``sigmoid", ``tanh", and rectified linear unit (ReLU) have received more attention than the other activation functions recently. The ``sigmoid" is a common choice, which takes a real-value and squashes it to the range $[0,1]$. This activation function is normally used in the output layer, where a cross-entropy loss function is used for training a classification model. The ``tanh" and ``sigmoid" activation functions are defined as
\begin{equation}
tanh(x)=\frac{e^{2x}-1}{e^{2x}+1}
\end{equation}
and 
\begin{equation}
\sigma(x)=\frac{1}{1+e^{-x}},
\end{equation}
respectively. The ``tanh" activation function is in fact a scaled ``sigmoid" activation function such as
\begin{equation}
\sigma(x)=\frac{tanh(x/2)+1}{2}.
\end{equation}
ReLU is another popular activation function, which is open-ended for positive input values \cite{bengio2013advances}, defined as
 \begin{equation}
 y(x) = max(x, 0).
 \end{equation}

Selection of the activation function is mostly dependent on the problem and nature of the data. For example, ``sigmoid" is suitable for networks where the output is in the range $[0,1]$. However, the ``tanh" and ``sigmoid" activation functions saturate the neuron very fast and can vanish the gradient. Despite ``tanh", the non-zero centered output from ``sigmoid" can cause unstable dynamics in the gradient updates for the weights. The ReLU activation function leads to sparser gradients and greatly accelerates the convergence of stochastic gradient descent (SGD) compared to the ``sigmoid" or ``tanh" activation functions \cite{krizhevsky2012imagenet}. ReLU is computationally cheap, since it can be implemented by thresholding an activation value at zero. However, ReLU is not resistant against a large gradient flow and as the weight matrix grows, the neuron may remain inactive during training.

\subsection{Loss Function}
Loss function evaluates performance of the network by comparing the output $\textbf{y}_{t}$ with the corresponding target $\textbf{z}_{t}$ defined as
\begin{equation}
 \mathcal{L}(\textbf{y}, \textbf{z})=\sum_{t=1}^{T}\mathcal{L}_{t}(\textbf{y}_{t}, \textbf{z}_{t}),
 \label{eq:loss_function}
\end{equation} 
that is an overall summation of losses in each timestep~\cite{sutskever2013training}. Selection of the loss function is problem dependent. Some popular loss function are Euclidean distance and Hamming distance for forecasing of real-values and cross-entropy over probablity distribution of outputs for classification problems~\cite{Goodfellow-et-al-2016}.

\begin{table*}[]
\centering
\footnotesize
\caption{Comparing major gradient descent (GD) methods, where $N$ is number of nodes in the network and $O(\cdot)$ is per data point. More details in~\cite{williams1995gradient}.}
\begin{tabular}{|c|l|l|l|c|}
\hline
Method & \multicolumn{1}{c|}{Description} & \multicolumn{1}{c|}{Advantages} & \multicolumn{1}{c|}{Disadvantages} & \multicolumn{1}{c|}{\begin{tabular}[c]{@{}l@{}}  $O(\cdot)$ \end{tabular}}  \\ \hline

RTRL   & \begin{tabular}[c]{@{}l@{}}  Computing error gradient after obtaining \\gradients of the network states w.r.t \\weights at time $t$ in terms of those at \\time $t-1$\end{tabular} &   \begin{tabular}[c]{@{}l@{}}  - online updating of weights\\ - suitable for online adaption\\ property applications  \end{tabular}  &  \begin{tabular}[c]{@{}l@{}}  - large computational \\ complexity \end{tabular}  & $O(N^{4})$ \\ \hline

BPTT   & \begin{tabular}[c]{@{}l@{}} Unfolding time iterations into layers with\\ identical weights converts the recurrent\\ network into an equivalent feedforward\\ network, suitable for training with\\ back-propagation method. \end{tabular} & \begin{tabular}[c]{@{}l@{}} - computationally efficient\\ - suitable for offline training \end{tabular} & \begin{tabular}[c]{@{}l@{}} - not practical for\\ online training\end{tabular} & $O(N^{2})$ \\ \hline

FFP    & \begin{tabular}[c]{@{}l@{}}Recursive computing of boundary\\ conditions of back-propagated gradients\\ at time $t=1$. \end{tabular} & \begin{tabular}[c]{@{}l@{}}- on-line technique\\ - solving the gradient\\ recursion forward in time,\\ rather than backwards.\end{tabular} &  \begin{tabular}[c]{@{}l@{}} - more computational \\complexity than \\ BPTT method \end{tabular} & $O(N^{3})$\\ \hline

GF     & \begin{tabular}[c]{@{}l@{}} Computing the solution using the sought\\ error gradient based on the recursive\\ equations for the output gradients and a\\ dot product. \end{tabular} & \begin{tabular}[c]{@{}l@{}}- improving RTRL\\ computational complexity\\ - online method\end{tabular}                                                                        & \multicolumn{1}{c|}{\begin{tabular}[c]{@{}l@{}} - more computational \\complexity than \\ BPTT method \end{tabular} }& $O(N^{3})$  \\ \hline

BU     & \begin{tabular}[c]{@{}l@{}} Updating the weights every $O(N)$ data\\ points using some aspects of the RTRL \\and BTT methods.\end{tabular} & - online method& \multicolumn{1}{c|}{\begin{tabular}[c]{@{}l@{}} - more computational \\complexity than \\ BPTT method \end{tabular} }& $O(N^{3})$  \\ \hline

\end{tabular}
\label{T:gradient_methods}
\end{table*}

\section{Training Recurrent Neural Network}
\label{sec:trainingRNN}

Efficient training of a RNN is a major problem. The difficulty is in proper initialization of the weights in the network and the optimization algorithm to tune them in order to minimize the training loss. The relationship among network parameters and the dynamics of the hidden states through time causes instability \cite{bengio1994learning}. A glance at the proposed methods in the literature shows that the main focus is to reduce complexity of training algorithms, while accelerating the convergence. However, generally such algorithms take a large number of iterations to train the model. Some of the approaches for training RNNs are multi-grid random search, time-weighted pseudo-newton optimization, GD, extended Kalman filter (EKF)  \cite{puskorius1994neurocontrol}, Hessian-free, expectation maximization (EM) \cite{ma1998unified}, approximated Levenberg-Marquardt \cite{chan1999training}, and global optimization algorithms.
 In this section, we discuss some of these methods in details. A detailed comparison is available in \cite{ruder2016overview}.

 \subsection{Initialization}
Initialization of weights and biases in RNNs is critical. A general rule is to assign small values to the weights. A Gaussian draw with a standard deviation of 0.001 or 0.01 is a reasonable choice \cite{sutskever2013importance}, \cite{pascanu2013difficulty}. The biases are usually set to zero, but the output bias can also be set to a very small value \cite{sutskever2013importance}. However, the initialization of parameters is dependent on the task and properties of the input data such as dimensionality~\cite{sutskever2013importance}. Setting the initial weight using prior knowledge or in a semi-supervised fashion are other approaches \cite{bengio1994learning}. 

\subsection{Gradient-based Learning Methods}
Gradient descent (GD) is a simple and popular optimization method in deep learning. The basic idea is to adjust the weights of the model by finding the error function derivatives with respect to each member of the weight matrices in the model \cite{bengio1994learning}. To minimize total loss, GD changes each weight in proportion to the derivative of the error with respect to that weight, provided that the non-linear activation functions are differentiable. 
The GD is also known as batch GD, as it computes the gradient for the whole dataset in each optimization iteration to perform a single update as
\begin{equation}
\theta_{t+1}=\theta_{t}-\frac{\lambda}{U} \sum^{U}_{k=1}\frac{\partial \mathcal{L}_{k}}{\partial \theta}
\label{eq:GDupdating}
\end{equation}
where $U$ is size of training set, $\lambda$ is the learning rate, and $\theta$ is set of parameters. This approach is computationally expensive for very large datasets and is not suitable for online training (i.e., training the models as inputs arrive).

Since a RNN is a structure through time, we need to extend GD through time to train the network, called back-propagation through time (BPTT) \cite{werbos1990backpropagation}. However, computing error-derivatives through time is difficult \cite{le2015simple}. This is mostly due to the relationship among the parameters and the dynamics of the RNN, that is highly unstable and makes GD ineffective. Gradient-based algorithms have difficulty in capturing dependencies as the duration of dependencies increases \cite{bengio1994learning}. The derivatives of the loss function with respect to the weights only consider the distance between the current output and the corresponding target, without using the history information for weights updating \cite{perez2003kalman}.
RNNs cannot learn long-range temporal dependencies when GD is used for training \cite{bengio1994learning}. This is due to the exponential decay of gradient, as it is back-propagated through time, which is called the vanishing gradient problem. In another occasional situation, the back-propagated gradient can exponentially blow-up, which increases the variance of the gradients and results in very unstable learning situation, called the exploding gradient problem \cite{sutskever2011generating}. These challenges are discussed in this section. A comparison of major GD methods is presented in Table~\ref{T:gradient_methods} and an overview of gradient-based optimization algorithms is provided in \cite{ruder2016overview}.

\begin{figure}[!t]
\footnotesize
\centering
\includegraphics[width=0.7\linewidth]{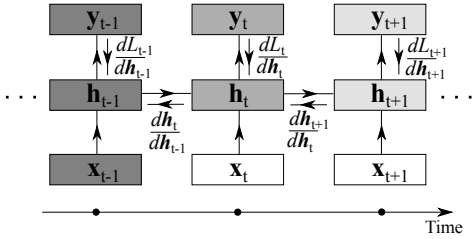}
\caption{As the network is receiving new inputs over time, the sensitivity of units decay (lighter shades in layers) and the back-propagation through time (BPTT) overwrites the activation in hidden units. This results in forgetting the early visited inputs.}
\label{Fig:bptt}     
\end{figure}

\subsubsection{Back-propagation through time (BPTT)}

BPTT is a generalization of back-propagation for feed-forward networks. The standard BPTT method for learning RNNs ``unfolds'' the network in time and propagates error
signals backwards through time. By considering the network parameters in Figure~\ref{fig:srnn_unfolded} as the set $\theta=\{\mathbf{W}_{HH}, \mathbf{W}_{IH}, \mathbf{W}_{HO}, \mathbf{b}_{H}, \mathbf{b}_{I}, \mathbf{b}_{O}\}$ and $\textbf{h}_{t}$ as the hidden state of network at time $t$, we can write the gradients as
\begin{equation}
\frac{\partial \mathcal{L}}{\partial \theta}=\sum_{t=1}^{T} \frac{\partial \mathcal{L}_{t}}{\partial \theta}
\end{equation}
where the expansion of loss function gradients at time $t$ is
\begin{equation}
\frac{\partial \mathcal{L}_{t}}{\partial \theta}=\sum_{k=1}^{t}(\frac{\partial \mathcal{L}_{t}}{\partial \textbf{h}_{t}} \cdot \frac{\partial {\textbf{h}}_{t}}{\partial \textbf{h}_{k}}.\frac{\partial \textbf{h}_{k}^{+}}{\partial \theta}) 
\end{equation}
where $\frac{\partial \textbf{h}_{k}^{+}}{\partial \theta}$ is the partial derivative (i.e., ``immediate'' partial derivative). It describes how the parameters in the set $\theta$ affect the loss function at the previous timesteps (i.e., $k<t$). In order to transport the error through time from timestep $t$ back to timestep $k$ we can have
\begin{equation}
\frac{\partial {\textbf{h}}_{t}}{\partial \textbf{h}_{k}}=\prod_{i=k+1}^{t}\frac{\partial {\textbf{h}}_{i}}{\partial \textbf{h}_{i-1}}.
\label{eq:sigmaloss}
\end{equation}
We can consider Eq.~(\ref{eq:sigmaloss}) as a Jacobian matrix for the hidden state parameters in Eq.(\ref{eq:SRNN_hidden_state}) as
\begin{equation}
\prod_{i=k+1}^{t}\frac{\partial {\textbf{h}}_{i}}{\partial \textbf{h}_{i-1}}=\prod_{i=k+1}^{t}\textbf{W}_{HH}^{T}diag|f^{'}_{H}(\textbf{h}_{i-1})|,
\label{jacob}
\end{equation}
where $f^{'}(\cdot)$ is the element-wise derivate of function $f(\cdot)$ and $diag(\cdot)$ is the diagonal matrix. 

We can generally recognize the long-term and short-term contribution of hidden states over time in the network. The long-term dependency refers to the contribution of inputs and corresponding hidden states at time $k<<t$ and short-term dependencies refer to other times \cite{pascanu2013difficulty}. Figure~\ref{Fig:bptt} shows that as the network makes progress over time, the contribution of the inputs $x_{t-1}$ at discrete time $t-1$ vanishes through time to the timestep $t+1$ (the dark grey in the layers decays to higher grey). On the other hand, the contribution of the loss function value $\mathcal{L}_{t+1}$ with respect to the hidden state $\textbf{h}_{t+1}$ at time $t+1$ in BPTT is more than the previous timesteps.  

\subsubsection{Vanishing Gradient Problem}
According to the literature, it is possible to capture complex patterns of data in real-world by using a strong nonlinearity \cite{mikolov2014learning}. However, this may cause RNNs to suffer from the vanishing gradient problem \cite{bengio1994learning}. This problem refers to the exponential shrinking of gradient magnitudes as they are propagated back through time. This phenomena causes memory of the network to ignore long term dependencies and hardly learn the correlation between temporally distant events. There are two reasons for that: 1)~Standard nonlinear functions such as the sigmoid function have a gradient which is almost everywhere close to zero; 2)~The magnitude of gradient is multiplied over and over by the recurrent matrix as it is back-propagated through time. In this case, when the eigenvalues of the recurrent matrix become less than one, the gradient converges to zero rapidly. This happens normally after 5$\sim$10 steps of back-propagation \cite{mikolov2014learning}.  

In training the RNNs on long sequences (e.g., 100 timesteps), the gradients shrink when the weights are small. Product of a set of real numbers can shrink/explode to zero/infinity, respectively. For the matrices the same analogy exists but shrinkage/explosion happens along some directions. In \cite{pascanu2013difficulty}, it is showed that by considering $\rho$ as the spectral radius of the recurrent weight matrix $W_{HH}$, it is necessary at $\rho>1$ for the long term components to explode as $t\rightarrow \infty$.
It is possible to use singular values to generalize it to the non-linear function $f^{'}_{H}(\cdot)$ in Eq. (\ref{eq:SRNN_hidden_state}) by bounding it with $\gamma\in\mathcal{R}$ such as

\begin{equation}
||diag(f^{'}_{H}(\textbf{h}_{k}))||\leq \gamma.
\label{bound}
\end{equation}
Using the Eq. (\ref{jacob}), the Jacobian matrix $\frac{\partial {\textbf{h}}_{k+1}}{\partial \textbf{h}_{k}}$, and the bound in Eq. (\ref{bound}),  we can have
\begin{equation}
||\frac{\partial {\textbf{h}}_{k+1}}{\partial \textbf{h}_{k}}|| \leq ||\textbf{W}_{HH}^{T}||\cdot ||diag(f^{'}_{H}(\textbf{h}_{k})) ||\leq 1.
\end{equation}
We can consider $||\frac{\partial {\textbf{h}}_{k+1}}{\partial \textbf{h}_{k}}|| \leq \delta <1$ such as $\delta \in \mathcal{R}$ for each step $k$. By continuing it over different timesteps and adding the loss function component we can have
  
\begin{equation}
||\frac{\partial \mathcal{L}_{t}}{\partial \textbf{h}_{t}}  (\prod_{i=k}^{t-1}\frac{\partial {\textbf{h}}_{i+1}}{\partial \textbf{h}_{i}})|| \leq \delta^{t-k}||\frac{\partial \mathcal{L}_{t}}{\partial \textbf{h}_{t}} ||.
\end{equation}
 This equation shows that as $t-k$ gets larger, the long-term dependencies move toward zero and the vanishing problem happens. Finally, we can see that the sufficient condition for the gradient vanishing problem to appear is that the largest singular value of the recurrent weights matrix $\textbf{W}_{HH}$ (i.e., $\lambda_{1}$) satisfies $\lambda_{1} < \frac{1}{\gamma}$ \cite{pascanu2013difficulty}. 

\subsubsection{Exploding Gradient Problem}
 
One of the major problems in training RNNs using BPTT is the exploding gradient problem  \cite{bengio1994learning}. Gradients in training RNNs on long sequences may explode as the weights become larger and the norm of the gradient during training largely increases. As it is stated in \cite{pascanu2013difficulty}, the necessary condition for this situation to happen is $\lambda_{1}>\frac{1}{\gamma}$. 

In order to overcome the exploding gradient problem, many methods have been proposed recently. In 2012,  Mikolov proposed a gradient norm-clipping method to avoid the exploding gradient problem in training RNNs with simple tools such as BPTT and SGD on large datasets. \cite{mikolov2012subword}, \cite{mikolov2012context}. In a similar approach, Pascanu has proposed an almost similar method to Mikolov, by introducing a hyper-parameter as threshold for norm-clipping the gradients \cite{pascanu2013difficulty}. This parameter can be set by heuristics; however, the training procedure is not very sensitive to that and behaves well for rather small thresholds.  


\subsubsection{Stochastic Gradient Descent}
The SGD (also called on-line GD) is a generalization of GD that is widely in use for machine learning applications \cite{sutskever2013training}. The SGD is robust, scalable, and performs
well across many different domains ranging from smooth and strongly convex problems to complex
non-convex objectives. Despite the redundant computations in GD, the SGD performs one update at a time \cite{lecun2012efficient}. For an input-target pair $\{\mathbf{x}_{k}, \mathbf{z}_{}\}$ in which $k\in\{1,...,U\}$, the parameters in $\theta$ are updated according as
\begin{equation}
\theta_{t+1}=\theta_{t}-\lambda \frac{\partial \mathcal{L}_{k}}{\partial \theta}.
\label{eq:update}
\end{equation}
Such frequent update causes fluctuation in the loss function outputs, which helps the SGD to explore the problem landscape with higher diversity with the hope of finding better local minima. An adaptive learning rate can control the convergence of SGD, such that as learning rate decreases, the exploration decreases and exploitation increases. It leads to faster convergence to a local minima. 
\begin{figure}[!t]
	\footnotesize
	\centering
	\begin{subfigure}[b]{0.24\textwidth}
		\centering 
		\includegraphics[width=1\linewidth]{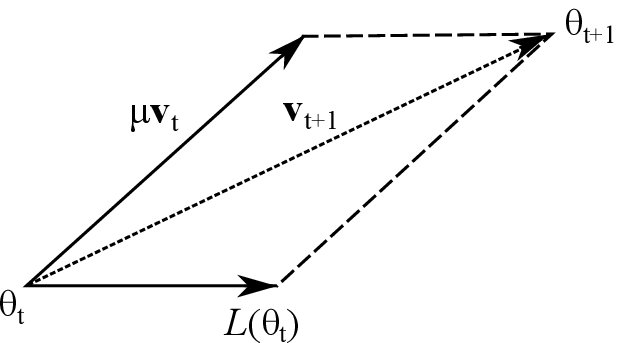}
		\caption{Classical momentum.}
		\label{fig:classical_momentum}
	\end{subfigure}%
	~
	\begin{subfigure}[b]{0.24\textwidth}
		\centering 
		\includegraphics[width=1\linewidth]{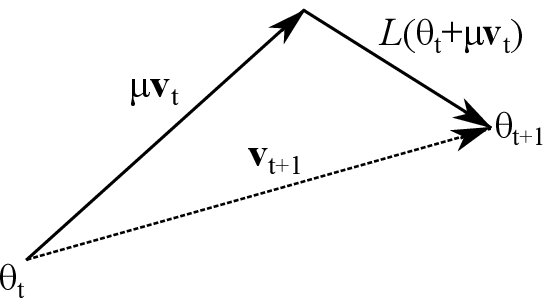}
		\caption{Nesterov accelerated gradient.}
		\label{fig:nes_momentum}
	\end{subfigure}%
	
	\caption{The classical momentum and the Nesterov accelerated gradient schemes.}
	\label{fig:momentum}     
\end{figure}
A classical technique to accelerate SGD is using momentum, which accumulates a velocity vector in directions of persistent reduction towards the objective across iterations \cite{polyak1964some}. The classical version of momentum applies to the loss function $\mathcal{L}$ at time $t$ with a set of parameters $\theta$ as
 
\begin{equation}
\textbf{v}_{t+1}=\mu\textbf{v}_{t}-\lambda \nabla\mathcal{L}(\theta_{t})
\end{equation}
where $\nabla\mathcal{L}(\cdot)$ is the gradient of loss function and $\mu\in[0,1]$ is the momentum coefficient \cite{sutskever2013importance}, \cite{sutskever2013training}. As figure~\ref{fig:classical_momentum} shows, the parameters in $\theta$ are updated as
\begin{equation}
\theta_{t+1}=\theta_{t}+\textbf{v}_{t+1}.
\label{eq:updateCM}
\end{equation}
By considering $R$ as the condition
number of the curvature at the minimum, the momentum can considerably accelerate
convergence to a local minimum, requiring $\sqrt{R}$
times fewer iterations than steepest descent to reach
the same level of accuracy \cite{polyak1964some}. In this case, it is suggested to set the learning rate to $\mu=(\sqrt{R}-1)/(\sqrt{R}+1)$ \cite{polyak1964some}.

The Nesterov accelerated gradient (NAG) is a first-order optimization method that provides more efficient convergence rate for particular situations (e.g., convex functions with deterministic gradient) than the GD \cite{cotter2011better}. The main difference between NAG and GD is in the updating rule of the velocity vector $\textbf{v}$, as presented in Figure~\ref{fig:nes_momentum}, defined as

\begin{equation}
\textbf{v}_{t+1}=\mu\textbf{v}_{t}-\lambda \nabla\mathcal{L}(\theta+\mu\textbf{v}_{t})
\end{equation}
where the parameters in $\theta$ are updated using Eq.~(\ref{eq:updateCM}). By reasonable fine-tuning of the momentum coefficient $\mu$, it is possible to increase the optimization performance \cite{sutskever2013importance}. 


\subsubsection{Mini-Batch Gradient Descent}
The mini-batch GD computes the gradient of a batch of training data which has more than one training sample. The typical mini-batch size is $50\leq b\leq256$, but can vary for different applications. Feeding the training samples in mini-batches accelerates the GD and is suitable for processing load distribution on graphical processing units (GPUs). The update rule modifies the parameters after $b$ examples rather than needing to wait to scan the all examples such as
\begin{equation}
\theta_{t}=\theta_{t-1}-\frac{\lambda}{b} \sum^{i+b-1}_{k=i}\frac{\partial \mathcal{L}_{k}}{\partial \theta}.
\end{equation}
Since the GD-based algorithms are generally dependent on instantaneous estimations
of the gradient, they are slow for time series data \cite{perez2003kalman} and ineffective on optimization of non-convex functions \cite{martens2012training}. They also require setting of learning rate which is often tricky and application dependent. 

The SGD is much faster than the GD and is useable for tracking of updates. However, since the mini-batch GD is easier for parallelization and can take advantage of vectorized implementation, it performs significantly better than GD and SGD \cite{lecun2012efficient}. A good vectorization can even lead to faster results compared to SGD. Also non-random initializations schemes, such as layer-wise pre-training, may help with faster optimization  \cite{bengio2007greedy}. A deeper analyze is provided in \cite{bottou2004stochastic}. 

\subsubsection{Adam Stochastic Optimization}
Adaptive Moment Estimation (Adam) is a first-order gradient-based optimization algorithm, which uses estimates of lower-order moments to optimize 
a stochastic objective function \cite{kingma2014adam}. It needs initialization of first moment vector $\textbf{m}_{0}$ and second moment vector $\textbf{v}_{0}$ at time-stamp zero. These vector are updated as
\begin{equation}
\textbf{m}_{t+1}=\beta_{1}\textbf{m}_{t}+(1-\beta_{1})g_{t+1}
\end{equation}
and 
\begin{equation}
\textbf{v}_{t+1}=\beta_{2}\textbf{v}_{t}+(1-\beta_{2})g_{t+1}^{2},
\end{equation}
where $g_{t+1}$ is the gradient of loss function. The exponential decay rates for the moment estimates are recommended to be $\beta_{1}=0.9 $ and $\beta_{2}=0.999$ \cite{kingma2014adam}. The bias correction of first and second moment estimates are 
\begin{equation}
\hat{\textbf{m}}_{t+1}=\hat{\textbf{n}}_{t+1}=\textbf{v}_{t}/(1-\beta_{1}^{t+1}),
\end{equation}
and 
\begin{equation}
\hat{\textbf{v}}_{t+1}=\textbf{v}_{t}/(1-\beta_{2}^{t+1}).
\end{equation}
Then, the parameters are updated as 
\begin{equation}
\theta_{t+1} = \theta_{t}-\frac{\alpha\cdot \hat{\textbf{m}}_{t+1}}{\sqrt{\hat{\textbf{v}}_{t}}+\epsilon}
\end{equation}
where $\epsilon=10^{-8}$. The Adam algorithm is relatively simple to implement and is suitable for problems with very large datasets~\cite{kingma2014adam}.

\subsection{Extended Kalman Filter-based Learning}
Kalman filter is a method of predicting the future state of a system based on a series of measurements observed over time by using Bayesian inference and estimating a joint probability distribution over the variables for each timestep \cite{haykin2001kalman}. The extended Kalman filter (EKF) is the nonlinear version of the Kalman filter. It relaxes the linear prerequisites of the state transition and observation models. However, they may instead need to be differentiable functions. 
The EKF trains RNNs with the assumption that the
optimum setting of the weights is stationary \cite{perez2003kalman}, \cite{williams1992training}.
Comparing to back-propagation, the EKF helps RNNs to reach the training steady state much faster
for non-stationary processes. It can excel the back-propagation algorithm in training with limited data \cite{puskorius1994neurocontrol}. Similar to SGD, it can train a RNN with incoming input data in an online manner~\cite{williams1992training}.

A more efficient and effective version of EKF is the decoupled EKF (DEKF) method, which ignores the interdependencies of mutually exclusive groups of weights \cite{haykin2001kalman}. This technique can lower the computational complexity and the required storage per training instance. 
The decoupled extended Kalman filter (DEKF) applies the extended Kalman
filter independently to each neuron in order to estimate the optimum weights feeding it. By
proceeding this way, only local interdependencies are considered.  The training procedure is modeled as an optimal filtering problem. It recursively and efficiently computes a solution
to the least-squares problem to find the best fitted curve for a given set of data in terms
of minimizing the average distance between data and curve. At a timestep $t$, all the
information supplied to the network until time $t$ is used, including all derivatives computed
since the first iteration of the learning process. However, computation requires just the results from the previous step and there is no need to store results beyond that step \cite{perez2003kalman}. Kalman-based models in RNNs are computationally expensive and have received little attention in the past years.

\subsection{Second Order Optimization}

The second order optimization algorithms use information of the second derivate of a function. With the assumption of having a quadratic function with good second order expansion approximation, Newton's method can perform better and faster than GD by moving toward the global minimum \cite{martens2010deep}. This is while the direction of optimization in GD is against the gradient and gets stuck near saddle points or local extrema. The other challenge with GD-based models is setting of learning rate, which is often tricky and application dependent. However, second order methods generally require computing the Hessian matrix and inverse of Hessian matrix, which is a difficult task to perform in RNNs comparing to GD approaches. 

A general recursive Bayesian Levenberg-Marquardt algorithm can sequentially update the weights and the Hessian matrix in recursive second-order training of a RNN \cite{mirikitani2010recursive}. Such approach outperforms standard real-time recurrent learning and EKF training algorithms for RNNs \cite{mirikitani2010recursive}. The challenges in computing Hessian matrix for time-series are addressed by introducing Hessian
free (HF) optimization \cite{martens2010deep}.

\subsection{Hessian-Free Optimization}
A well-designed and well-initialized HF optimizer can work very well for optimizing non-convex functions, such as training the objective function for deep neural networks, given sensible random initializations \cite{martens2010deep}. Since RNNs share weights across time, the HF optimizer should be a good optimization candidate \cite{sutskever2011generating}. Training
RNNs via HF optimization can
reduce training difficulties caused by gradient-based optimization~\cite{martens2011learning}. In general, HF and truncated Newton methods compute a new estimate
of the Hessian matrix before each update step and can
take into account abrupt changes in curvature~\cite{pascanu2013difficulty}. HF optimization targets unconstrained minimization of real-valued smooth objective functions \cite{martens2012training}. Like standard Newton's method, it uses local quadratic approximations to generate update proposals. It belongs to the broad class of approximate Newton methods that are practical for problems of very high dimensionality, such as the training objectives of large neural networks \cite{martens2012training}.

With the addition of a novel damping mechanism to a HF optimizer, the optimizer is able to train a RNN on pathological synthetic datasets, which are known to be impossible to learn with GD \cite{martens2012training}. Multiplicative RNNs (MRNNs) uses multiplicative (also called ``gated'') connections to allow the current input character to determine the transition matrix from one hidden state vector to the next~\cite{sutskever2011generating}. 
This method demonstrates the power of a large
RNN trained with this optimizer by applying
them to the task of predicting the next character in a
stream of text \cite{sutskever2011generating}, \cite{sutskever2013training}. 

The HF optimizer can be used in conjunction with or as an alternative to existing pre-training methods and is more widely applicable, since it relies on fewer assumptions about the specific structure of the network.
HF optimization operates on large mini batches and is able to detect promising directions in the weight space that have very small gradients but even smaller curvature. Similar results have been achieved by using SGD with momentum and initializing weights to small values close to zero~\cite{sutskever2013importance}. 


%
%
%

\subsection{Global Optimization}
In general, evolutionary computing methods initialize a population of search agents and evolve them to find local/global optimization solution(s)~\cite{salehinejad2017micro}. These methods can solve a wide range of optimization problems including multimodal, ill-behaved, high-dimensional, convex, and non-convex problems. However, evolutionary algorithms have some drawbacks in optimization of RNNs including getting stuck in local minima/maxima, slow speed of convergence, and network stagnancy.

Optimization of the parameters in RNNs can be modelled as a nonlinear global optimization problem. The most common global optimization method for training RNNs is genetic algorithms \cite{angeline1994evolutionary}. The Alopex-based evolutionary algorithm (AEA) uses local correlations between changes in individual weights and changes in the global error measure and simultaneously updates all the weights in the network using only local computations \cite{unnikrishnan1994alopex}. Selecting the optimal topology of neural network for a particular application is a different task from optimizing the network parameters. A hybrid multi-objective evolutionary algorithm that trains and optimizes the structure of a RNN for time series prediction is proposed in \cite{smith2014evolutionary}. Some models simultaneously acquire both the structure and weights for recurrent networks \cite{angeline1994evolutionary}. The covariance matrix adaptation evolution strategy (CMA-ES) is a global optimization method used for tuning the parameters of a RNN for language models~\cite{tanaka2016evolutionary}. Published literature on global optimization methods for RNNs is scattered and has not received much attention from the research community. This lack is mainly due to the computational complexity of these methods. However, the multi-agent philosophy of such methods in a low computational complexity manner, such as models with small population size~\cite{salehinejad2014micro}, may result in much better performance than SGD.

\section{Recurrent Neural Networks Architectures}
\label{sec:RNNArch}
This section aims to provide an overview on the different architectures of RNNs and discuss the nuances between these models. 

\subsection{Deep RNNs with Multi-Layer Perceptron}
\label{subsec:deeprnn}

Deep architectures of neural networks can represent a function exponentially more efficient than shallow architectures. While recurrent networks are inherently deep in time given each hidden state is a function of all previous hidden states~\cite{graves2013speech}, it has been shown that the internal computation is in fact quite shallow \cite{pascanu2013construct}. In \cite{pascanu2013construct}, it is argued that adding one or more nonlinear layers in the transition stages of a RNN can improve overall performance by better disentangling the underlying variations the original input. The deep structures in RNNs with perceptron layers can fall under three categories: input to hidden, hidden to hidden, and hidden to output \cite{pascanu2013construct}. 

\subsubsection{Deep input to hidden}
One of the basic ideas is to bring the multi-layer perceptron (MLP) structure into the transition and output stages, called deep transition RNNs and deep output RNNs, respectively. To do so, two operators can be introduced. The first is a plus~$\oplus$~operator, which receives two vectors, the input vector $\mathbf{x}$ and hidden state $\mathbf{h}$, and returns a summary as
\begin{equation}
\mathbf{h}^{'}=\mathbf{x}\oplus \mathbf{h}. 
\end{equation}
This operator is equivalent to the Eq.~(\ref{eq:SRNN_hidden_state}). The other operator is a predictor denoted as $\rhd$, which is equivalent to the Eq.~(\ref{eq:SRNN_outcome}) and predicts the output of a given summary $\mathbf{h}$ as
\begin{equation}
\mathbf{y}=\rhd \mathbf{h}. 
\end{equation}
Higher level representation of input data means easier representation of relationships between temporal structures of data. This technique has achieved better results than feeding the network with original data in speech recognition \cite{graves2013speech} and word embedding \cite{mikolov2013distributed} applications. Structure of a RNN with an MLP in the input to hidden layers is presented in Figure~\ref{fig:deep_in_h1}. In order to enhance long-term dependencies, an additional connection makes a short-cut between the input and hidden layer as in Figure~\ref{fig:deep_in_h2}~\cite{pascanu2013construct}.

\subsubsection{Deep hidden to hidden and output}
The most focus for deep RNNs is in the hidden layers. In this level, the procedure of data abstraction and/or hidden state construction from previous data abstractions and new inputs is highly non-linear. An MLP can model this non-linear function, which helps a RNN to quickly adapt to fast changing input modes while still having a good memory of past events.
A RNN can have both an MLP in transition and an MLP before the output layer (an example is presented in Figure~\ref{fig:deep_in_h3})~\cite{pascanu2013construct}. A deep hidden to output function can disentangle the factors of variations in the hidden state and facilitate prediction of the target. This function allows a more compact hidden state of the network, which may result in a more informative historical summary of the previous inputs.

\begin{figure}
	\footnotesize
	\centering
	\begin{subfigure}[b]{0.22\textwidth}
		\centering 
		\includegraphics[width=0.8\linewidth]{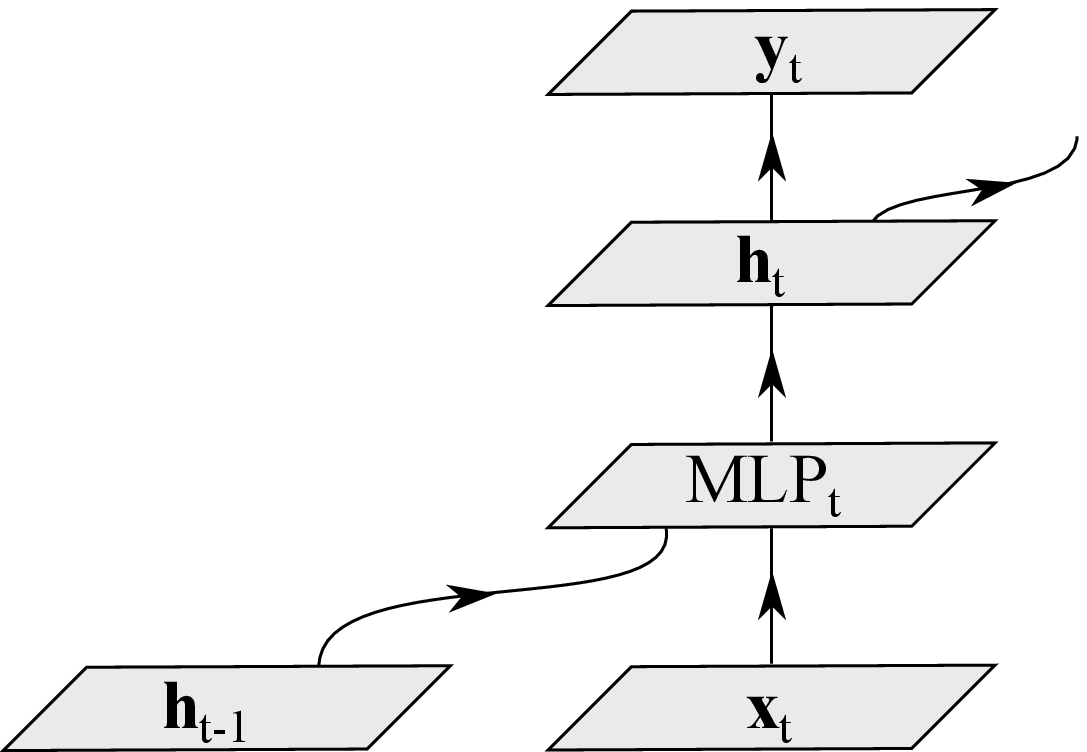}
		\caption{Input to hidden.}
		\label{fig:deep_in_h1}
	\end{subfigure}%
	~
	\begin{subfigure}[b]{0.22\textwidth}
		\centering 
		\includegraphics[width=0.8\linewidth]{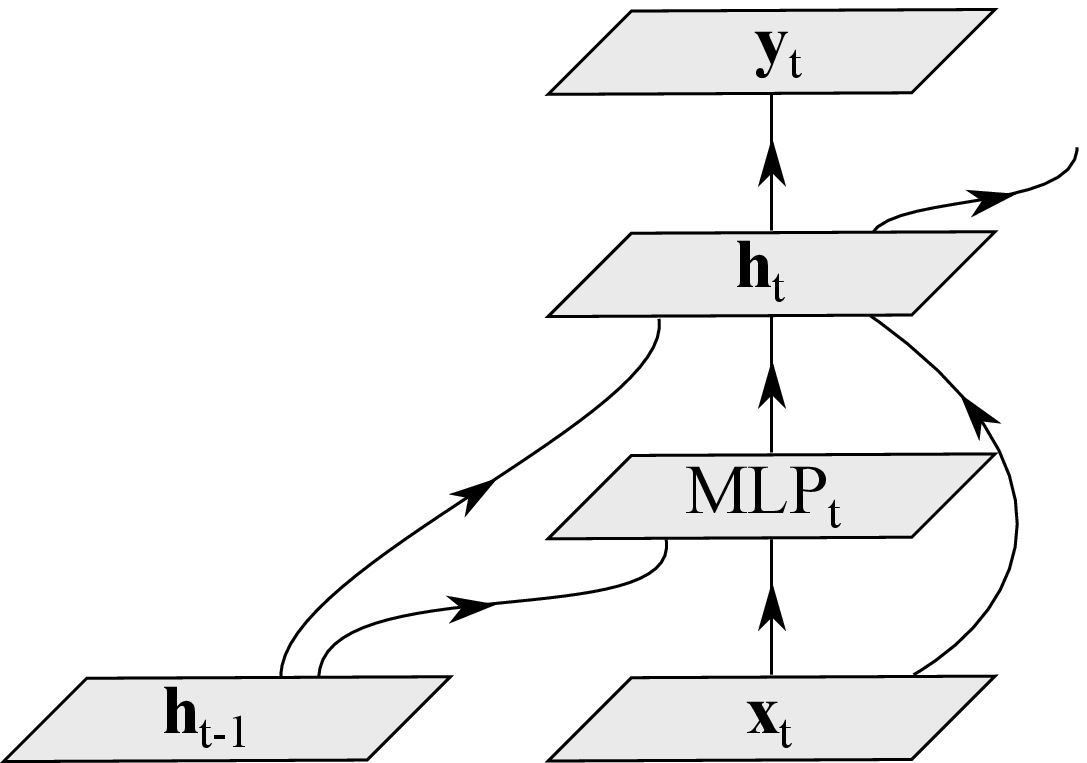}
		\caption{Input to hidden with short-cut.}
		\label{fig:deep_in_h2}
	\end{subfigure}%
	
	\begin{subfigure}[b]{0.22\textwidth}
		\centering 
		\includegraphics[width=0.8\linewidth]{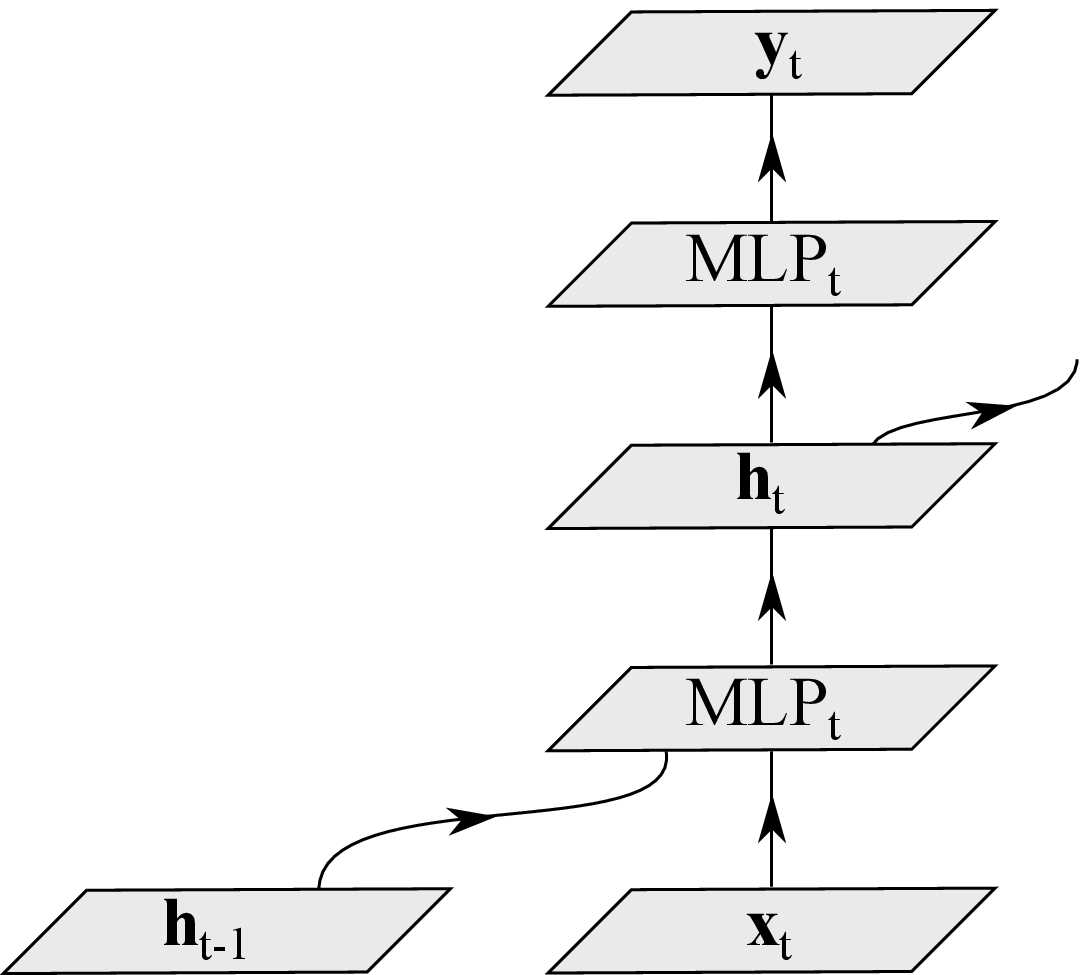}
		\caption{Hidden to hidden and output.}
		\label{fig:deep_in_h3}
	\end{subfigure}%
	~
	\begin{subfigure}[b]{0.22\textwidth}
		\centering 
		\includegraphics[width=0.8\linewidth]{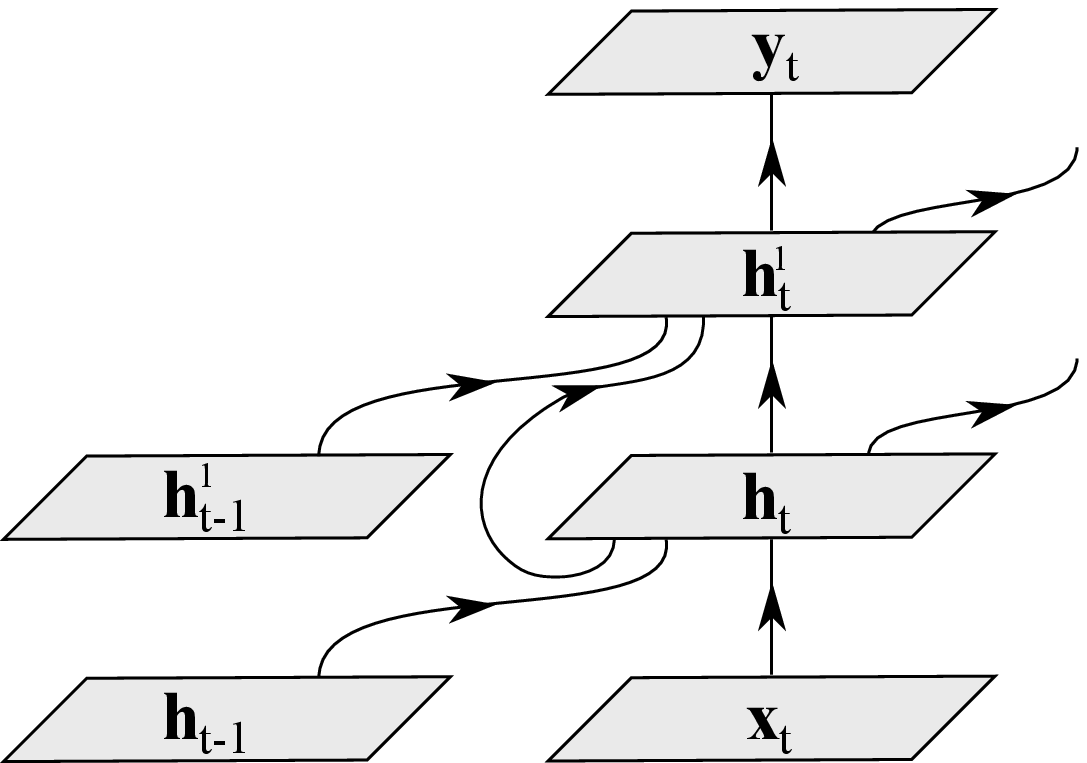}
		\caption{Stack of hidden states.}
		\label{fig:deep_in_h4}
	\end{subfigure}%
	
	\caption{Some deep recurrent neural network (RNN) architectures with multi-layer perceptron (MLP).}
	\label{fig:deep_rnn}     
\end{figure}

\subsubsection{Stack of hidden states}
Another approach to construct deep RNNs is to have a stack of hidden recurrent layers as shown in Figure~\ref{fig:deep_in_h4}. This style of recurrent levels encourages the network to operate at different timescales and enables it to deal with multiple time scales of inputs sequences~\cite{pascanu2013construct}. However, the transitions between consecutive hidden states is often shallow, which results in a limited family of functions it can represent~\cite{pascanu2013construct}. Therefore, this function cannot act as a universal approximation, unless the higher layers have feedback to the lower layers.

While the augmentation of a RNN for leveraging the benefits of deep networks has shown to yield performance improvements, it has also shown to introduce potential issues. By adding nonlinear layers to the network transition stages, there now exists additional layers through which the gradient must travel back. This can lead to issues such as vanishing and exploding gradients which can cause the network to fail to adequately capture long-term dependencies~\cite{pascanu2013construct}. The addition of nonlinear layers in the transition stages of a RNN can also significantly increase the computation and speed of the network. Additional layers can significantly increase the training time of the network, must be unrolled at each iteration of training, and can thus not be parallelized. 

\begin{figure}[!t]
\footnotesize

\centering
\includegraphics[width=0.7\linewidth]{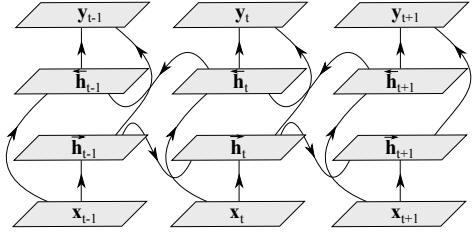}

\caption{Unfolded through time bi-directional recurrent neural network (BRNN).}
\label{fig:BRNN}
\end{figure}

\subsection{Bidirectional RNN}
\label{subsec:BRNN}
Conventional RNNs only consider the previous context of data for training. While simply looking at previous context is sufficient in many applications such as speech recognition, it is also useful to explore the future context as well \cite{graves2013speech}. Previously, the use of future information as context for current prediction have been attempted in the basic architecture of RNNs by delaying the output by a certain number of time frames. However, this method required a handpicked optimal delay to be chosen for any implementation. A bi-directional RNN (BRNN) considers all available input sequence in both the past and future for estimation of the output vector \cite{schuster1997bidirectional}. To do so, one RNN processes the sequence from start to end in a forward time direction. Another RNN processes the sequence backwards from end to start in a negative time direction as demonstrated in Figure~\ref{fig:BRNN}. Outputs from forward states are not connected to inputs of backward states and vice versa and there are no interactions between the two types of state
neurons~\cite{schuster1997bidirectional}. In Figure~\ref{fig:BRNN}, the forward and backward hidden sequences are denoted by $\overset{\rightarrow}{\mathbf{h}}_{t}$ and $\overset{\leftarrow}{\mathbf{h}}_{t}$, respectively, at time $t$. The forward  hidden sequence is computed as
\begin{equation}
\overset{\rightarrow}{\mathbf{h}_{t}} = f_{H}(\mathbf{W}_{\overset{\rightarrow}{IH}}\mathbf{x}_{t}+\mathbf{W}_{\overset{\rightarrow}{HH}}\overset{\rightarrow}{\mathbf{h}}_{t-1}+\mathbf{b}_{\overset{\rightarrow}{\mathbf{h}}}),
\label{eq:BRNN_hidden_state_f}
\end{equation}
where it is iterated over $t=(1,...,T)$. The backward layer is
\begin{equation}
\overset{\leftarrow}{\mathbf{h}_{t}} = f_{H}(\mathbf{W}_{\overset{\leftarrow}{IH}}\mathbf{x}_{t}+\mathbf{W}_{\overset{\leftarrow}{HH}}\overset{\leftarrow}{\mathbf{h}}_{t-1}+\mathbf{b}_{\overset{\leftarrow}{\mathbf{h}}}),
\label{eq:BRNN_hidden_state_b}
\end{equation}
which is iterated backward over time $t=(T,...,1)$. The output sequence $\mathbf{y}_{t}$ at time $t$ is
\begin{equation}
\mathbf{y}_{t} = \mathbf{W}_{\overset{\rightarrow}{HO}}\overset{\rightarrow}{\mathbf{h}_{t}} + \mathbf{W}_{\overset{\leftarrow}{HO}}\overset{\leftarrow}{\mathbf{h}_{t}} + \mathbf{b}_{o}.
\label{eq:BRNN_outcome}
\end{equation}

BPTT is one option to train BRNNs. However, the forward and backward pass procedures are slightly more complicated because the update of state and output neurons can no longer be conducted one at a time \cite{schuster1997bidirectional}. While simple RNNs are constrained by inputs leading to the present time, the BRNNs extend this model by using both past and future information. However, the shortcoming of BRNNs is their requirement to know the start and end of input sequences in advance. An example is labeling spoken sentences by their phonemes \cite{schuster1997bidirectional}.

\begin{figure}[!t]
\footnotesize
\centering
\begin{subfigure}[b]{0.49\textwidth}
           \centering
                 \includegraphics[width=0.7\linewidth]{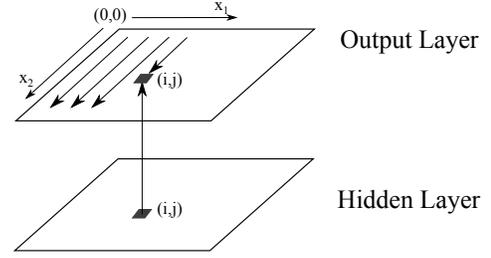}
        \end{subfigure}%

\caption{Forward pass with sequence ordering in two-dimensional recurrent neural network (RNN). 
The connections within the hidden layer plane are recurrent. The lines along $x_{1}$ and $x_{2}$ show the scanning
strips along which previous points were visited, starting at the top left corner.}
\label{fig:mdrnn_ff2}
\end{figure}
\subsection{Recurrent Convolutional Neural Networks}
The rise in popularity of RNNs can be attributed to its ability to model sequential data. Previous models examined have augmented the underlying structure of a simple RNN to improve its performance on learning the contextual dependencies of single dimension sequences. However, there exists several problems, which require understanding of contextual dependencies over multiple dimensions. The most popular network architectures use convolutional neural networks (CNNs) to tackle these problems. 

CNNs are very popular models for machine vision applications. CNNs may consist of multiple convolutional layers, optionally with pooling layers in between, followed by fully connected perceptron layers \cite{krizhevsky2012imagenet}. Typical CNNs learn through the use of convolutional layers to extract features using shared weights in each layer. The feature pooling layer (i.e., sub-sampling) generalizes the network by reducing the resolution of the dimensionality of intermediate representations (i.e., feature maps) as well as the sensitivity of the output to shifts and distortions. The extracted features, at the very last convolutional layer, are fed to fully connected perceptron model for dimensionality reduction of features and classification. 


Incorporation of recurrent connections into each convolutional layer can shape a recurrent convolutional neural network (RCNN) \cite{liang2015recurrent}. The activation of units in RCNN evolve over time, as they are dependent on the neighboring unit. This approach can integrate the context information, important for object recognition tasks. This approach increases the depth of model, while the number of parameters is constant by weight sharing between layers. 
%
Using recurrent connections from the output into the input of the hidden layer allows the network to model label dependencies and smooth its own outputs based on its previous outputs \cite{pinheiro2014recurrent}. This RCNN approach allows a large input context to be fed to the network while limiting the capacity of the model. This system can model complex spatial dependencies with low inference cost. As the context size increases with the built-in recurrence, the system identifies and corrects its own errors \cite{pinheiro2014recurrent}. Quad-directional 2-dimensional RNNs can enhance CNNs to model long range spatial dependencies~\cite{shuai2015quaddirectional}. This method efficiently embeds the global spatial context into the compact local representation~\cite{shuai2015quaddirectional}. 


\subsection{Multi-Dimensional Recurrent Neural Networks}
Multi-dimensional recurrent neural networks (MDRNNs) are another implementation of RNNs to high dimensional sequence learning. This network utilizes recurrent connections for each dimension to learn correlations in the data.  
MDRNNs are a special case of directed acyclic graph RNNs \cite{baldi2003principled}, generalized to multidimensional data by replacing the $one$-dimensional chain of network updates with a $D$-dimensional grid \cite{graves2007}. In this approach, the single recurrent connection is replaced with recurrent connections of size $D$. A 2-dimensional example is presented in Figure~\ref{fig:mdrnn_ff2}. During the forward pass at each timestep, the hidden layer receives an external input as well as its own activation from one step back along all dimensions. A combination of the input and the previous hidden layer activation at each timestep is fed in the order of input sequence. Then, the network stores the resulting hidden layer activation \cite{graves2009offline}. The error gradient of an MDRNN can be calculated with BPTT. As with one dimensional BPTT, the sequence is processed in the reverse order of the forward pass. At each timestep, the hidden layer receives both the output error derivatives and its own future derivatives \cite{graves2009offline}. 

RNNs have suitable properties for multidimensional domains such as robustness to warping and flexible use of context. Furthermore, RNNs can also leverage inherent sequential patterns in image analysis and video processing that are often ignored by other architectures \cite{visin2015renet}. However, memory usage can become a significant problem when trying to model multidimensional sequences. As more recurrent connections in the network are increased, so too is the amount of saved states that the network must conserve. This can result in huge memory requirements, if there is a large number of saved states in the network. MDRNNs also fall victim to vanishing gradients and can fail to learn long-term sequential information along all dimensions. 
While applications of the MDRNN fall in line with RCNNs, there has yet to be any comparative examinations performed on the two models. 

\subsection{Long-Short Term Memory}
Recurrent connections can improve performance of neural networks by leveraging their ability to understand sequential dependencies. However, the memory produced from the recurrent connections can severely be limited by the algorithms employed for training RNNs. All the models thus far have fallen victim to exploding or vanishing gradients during the training phase, resulting in the network failing to learn long-term sequential dependencies in data. The following models are specifically designed to tackle this problem, the most popular being the long-short term memory (LSTM) RNNs. 

LSTM is one of the most popular and efficient methods for reducing the effects of vanishing and exploding gradients~\cite{hochreiter1997long}. This approach changes the structure of hidden units from ``sigmoid" or ``tanh" to memory cells, in which their inputs and outputs are controlled by gates. These gates control flow of information to hidden neurons and preserve extracted features from previous timesteps~\cite{le2015simple}, \cite{hochreiter1997long}. 

It is shown that for a continual sequence, the LSTM model's internal values may grow without bound~\cite{gers1999learning}. Even when continuous sequences have naturally reoccurring properties, the network has no way to detect which information is no longer relevant. The forget gate learns weights that control the rate at which the value stored in the memory cell decays~\cite{gers1999learning}. For periods when the input and output gates are off and the forget gate is not causing decay, a memory cell simply holds its value over time so that the gradient of the error stays constant during back-propagation over those periods~\cite{le2015simple}. This structure allows the network to potentially remember information for longer periods.

LSTM suffers from high complexity in the hidden layer. For identical size of hidden layers, a typical LSTM has about four times more parameters than a simple RNN \cite{mikolov2014learning}. The objective at the time of proposing the LSTM method was to introduce a scheme that could improve learning long-range dependencies, rather than to find the minimal or optimal scheme~\cite{le2015simple}. Multi-dimensional and grid LSTM networks have partially enhanced learning of long-term dependencies comparing to simple LSTM, which are discussed in this section.



\begin{figure}[!t]
\footnotesize
\centering
\includegraphics[width=0.6\linewidth]{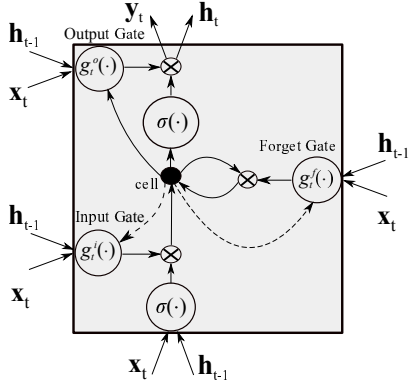}
\caption{The LSTM memory block with one cell. The dashed line represent time lag.}
\label{Fig:lstm_block}
\end{figure}

\subsubsection{Standard LSTM}
A typical LSTM cell is made of input, forget, and output gates and a cell activation component as shown in Figure~\ref{Fig:lstm_block}. These units receive the activation signals from different sources and control the activation of the cell by the designed multipliers. The LSTM gates can prevent the rest of the network from modifying the contents of the memory cells for multiple timesteps. LSTM networks preserve signals and propagate errors for much longer than ordinary RNNs. These properties allow LSTM networks to process data with complex and separated interdependencies and to excel in a range of sequence learning domains.


The input gate of LSTM is defined as
\begin{equation}
\mathbf{g}^{i}_{t}=\sigma(\mathbf{W}_{Ig^{i}}\mathbf{x}_{t} + \mathbf{W}_{Hg^{i}}\mathbf{h}_{t-1} + \mathbf{W}_{g^{c}g^{i}}  \mathbf{g}_{t-1}^{c} + \textbf{b}_{g^{i}}),
\label{eq:ggate}
\end{equation}
where $\mathbf{W}_{Ig^{i}}$ is the weight matrix from the input layer to the input gate, $\mathbf{W}_{Hg^{i}}$ is the weight matrix from hidden state to the input gate, $\mathbf{W}_{g^{c}g^{i}}$ is the weight matrix from cell activation to the input gate, and $b_{g^{i}}$ is the bias of the input gate. The forget gate is defined as
\begin{equation}
\mathbf{g}^{f}_{t}=\sigma(\mathbf{W}_{Ig^{f}}\mathbf{x}_{t} + \mathbf{W}_{Hg^{f}}\mathbf{h}_{t-1} + \mathbf{W}_{g^{c}g^{f}}  \mathbf{g}_{t-1}^{c} + \textbf{b}_{g^{f}}),
\label{eq:fgate}
\end{equation}
where $\mathbf{W}_{Ig^{f}}$ is the weight matrix from the input layer to the forget gate, $\mathbf{W}_{Hg^{f}}$ is the weight matrix from hidden state to the forget gate, $\mathbf{W}_{g^{c}g^{f}}$ is the weight matrix from cell activation to the forget gate, and $b_{g^{f}}$ is the bias of the forget gate. The cell gate is defined as
\begin{equation}
\mathbf{g}^{c}_{t}= \mathbf{g}_{t}^{i}~tanh(\mathbf{W}_{Ig^{c}}\mathbf{x}_{t} + \mathbf{W}_{Hg^{c}}\mathbf{h}_{t-1} + \textbf{b}_{g^{c}}) + \mathbf{g}_{t}^{f} \mathbf{g}_{t-1}^{c},
\label{eq:cgate}
\end{equation}
where $\mathbf{W}_{Ig^{c}}$ is the weight matrix from the input layer to the cell gate, $\mathbf{W}_{Hg^{c}}$ is the weight matrix from hidden state to the cell gate, and $b_{g^{c}}$ is the bias of the cell gate. The output gate is defined as
\begin{equation}
\mathbf{g}^{o}_{t}=\sigma(\mathbf{W}_{Ig^{o}}\mathbf{x}_{t} + \mathbf{W}_{Hg^{o}}\mathbf{h}_{t-1} + \mathbf{W}_{g^{c}g^{o}}  \mathbf{g}_{t}^{c} + \textbf{b}_{g^{o}}),
\label{eq:ogate}
\end{equation}
where $\mathbf{W}_{Ig^{o}}$ is the weight matrix from the input layer to the output gate, $\mathbf{W}_{Hg^{o}}$ is the weight matrix from hidden state to the output gate, $\mathbf{W}_{g^{c}g^{o}}$ is the weight matrix from cell activation to the output gate, and $\textbf{b}_{g^{o}}$ is the bias of the output gate.
Finally, the hidden state is computed as
\begin{equation}
\mathbf{h}_{t}=\mathbf{g}_{t}^{o}~tanh(\mathbf{g}_{t}^{c}).
\label{eq:hgate}
\end{equation}

\begin{figure}[!t]
\footnotesize
\centering
\includegraphics[width=0.5\linewidth]{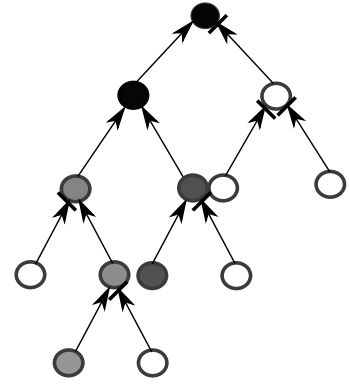}
\caption{An example of S-LSTM, a long-short term memory network
on tree structures. A tree node can consider information
from multiple descendants. Information of the other nodes in
white are blocked. The short line (-) at
each arrowhead indicates a block of information.}
\label{fig:slstm_tree}
\end{figure}

\subsubsection{S-LSTM}
While the LSTM internal mechanics help the network to learn longer sequence correlation, it may fail to understand input structures more complicated than a sequence. The S-LSTM model is designed to overcome the gradient vanishing problem and learn longer term dependencies from input. An S-LSTM network is made of S-LSTM memory blocks and works based on a hierarchical structure. A typical memory block is made of input and output gates. In this tree structure, presented in Figure~\ref{fig:slstm_tree}, the memory of multiple descendant cells over time periods are reflected on a memory cell recursively. This method learns long term dependencies over the input by considering information from long-distances on the tree (i.e., branches) to the principal (i.e., root). A typical S-LSTM has ``sigmoid" function and therefore, the gating signal works in the range of [0,1]. Figure~\ref{fig:slstm_tree} shows that the closer gates to the root suffer less from gradient vanishing problem (darker circle) while the branches at lower levels of tree loose their memory due to gradient vanishing (lighter circles). A gate can be closed to not receive signal from lower branches using a dash.

The S-LSTM method can achieve competitive results comparing to the recursive and LSTM models. It has the potential of extension to other LSTM models. However, its performance is not compared with other state-of-the-art LSTM models. The reader may refer to \cite{zhu2015long} for more details about S-LSTM memory cell.

\subsubsection{Stacked LSTM}
The idea of depth in ANNs is also applicable to LSTMs by stacking different hidden layers with LSTM cells in space to increase the network capacity~\cite{graves2013speech}, \cite{kalchbrenner2015grid}. A hidden layer $l$ in a stack of $L$ LSTMs using the hidden layer in Eq. (\ref{eq:SRNN_hidden_state}) is defined as
\begin{equation}
\textbf{h}_{t}^{l} = f_{H}(\textbf{W}_{IH}\textbf{h}_{t}^{l-1}+\textbf{W}_{HH}\textbf{h}_{t-1}^{l}+\textbf{b}_{h}^{l}),
\label{eq:StRNN_hidden_state}
\end{equation}
where the hidden vector sequence $\mathbf{h}^{l}_{t}$  is computed over time $t=(1,...,T)$ for $l=(1,...,L)$. The initial hidden vector sequence is defined using the input sequence~$\mathbf{h}^{0}=(\mathbf{x}_{1},...,\mathbf{x}_{T})$~\cite{graves2013speech}. Then, the output of the network is
\begin{equation}
\textbf{y}_{t} = f_{O}(\textbf{W}_{HO}\textbf{h}_{t}^{L}+\textbf{b}_{0}).
\label{eq:StRNN_outcome}
\end{equation}
In stacked LSTM, a stack pointer can determine which cell in the LSTM provides state and memory cell of a previous timestep \cite{yao2015depth}. In such a controlled structure, not only the controller can push to and pop from the top of the stack in constant time but also an LSTM can maintain a continuous space embedding of the stack contents~\cite{yao2015depth}, \cite{Ballesteros2015}.

The combination of stacked LSTM with different RNN structures for different applications needs investigation. One example is combination of stacked LSTM with frequency domain CNN for speech processing~\cite{graves2013speech}, \cite{Abdel2012}.

\subsubsection{Bidirectional LSTM}

It is possible to increase capacity of BRNNs by stacking hidden layers of LSTM cells in space, called deep bidirectional LSTM (BLSTM)~\cite{graves2013speech}. BLSTM networks are more powerful than unidirectional LSTM networks~\cite{graves2005framewise}. These networks theoretically involve all information of input sequences during computation. The distributed representation feature of BLSTM is crucial for different applications such as language understanding \cite{wang2015}. The BLSTM model leverages the same advantages discussed in the Bidirectional RNN section, while also overcoming the the vanishing gradient problem. 

\subsubsection{Multidimensional LSTM}

The classical LSTM model has a single self-connection which is controlled by a single forget gate. Its activation is considered as one dimensional LSTM. Multi-dimensional LSTM (MDLSTM) uses interconnection from previous state of cell to extend the memory of LSTM along every $N$ dimensions \cite{graves2009offline}, \cite{graves2007multi}. The MDLSTM receives inputs in a $N$-dimensional arrangement (e.g. two dimensions for an image).
Hidden state vectors $(\mathbf{h}_{1},...,\mathbf{h}_{N})$ and memory vectors $(\mathbf{m}_{1},...,\mathbf{m}_{N})$ are fed to each input of the array from the previous state for each dimension.  The memory vector is defined as
\begin{equation}
\mathbf{m}=\sum_{j=1}^{N} \mathbf{g}_{j}^{f} \odot \mathbf{m}_{j}+\mathbf{g}^{i}_{j}\odot\mathbf{g}^{c}_{j},
\end{equation}
where $\odot$ is the element-wise product and the gates are computed using Eq.(\ref{eq:ggate}) to Eq.(\ref{eq:hgate}), \cite{kalchbrenner2015grid}.

Spatial LSTM is a particular case of MDLSTM \cite{theis2015}, which is a two-dimensional grid for image modelling. This model generates a hidden state vector for a particular pixel in an image by sequentially reading the pixels in its small neighbourhood \cite{theis2015}. The state of the pixel is generated by feeding the state hidden vector into a factorized mixture of conditional Gaussian scale mixtures (MCGSM) \cite{theis2015}. 

\subsubsection{Grid LSTM}
The MDLSTM model becomes unstable, as the grid size and LSTM depth in space grows. The grid LSTM model provides a solution by altering the computation of output memory vectors. This method targets deep sequential computation of multi-dimensional data. The model connects LSTM cells along the spatiotemporal dimensions of input data and between the layers. Unlike the MDLSTM model, the block computes $N$ transforms and outputs $N$ hidden state vectors and $N$ memory vectors. The hidden sate vector for dimension $j$ is
\begin{equation}
\mathbf{h}^{'}_{j}=LSTM(\mathbf{H},\mathbf{m}_{j},\mathbf{W}_{{g}^{i}}^{j},\mathbf{W}_{{g}^{f}}^{j},\mathbf{W}_{{g}^{i}}^{o},\mathbf{W}_{{g}^{c}}^{j}),
\end{equation}
where $LSTM(\cdot)$ is the standard LSTM procedure \cite{kalchbrenner2015grid} and $\mathbf{H}$ is concatenation of input hidden state vectors defined as 
\begin{equation}
\mathbf{H}=[\mathbf{h}_{1},...,\mathbf{h}_{N}]^{T}.
\end{equation}

A two-dimension grid LSTM network adds LSTM cells along the spatial dimension to a stacked LSTM. A three or more dimensional LSTM is similar to MSLSTM, however, has added LSTM cells along the spatial depth and performs $N$-way interaction. More details on grid LSTM are provided in \cite{kalchbrenner2015grid}.


\subsubsection{Differential Recurrent Neural Networks}
\begin{figure}[!t]
\footnotesize
\centering
\includegraphics[width=0.6\linewidth]{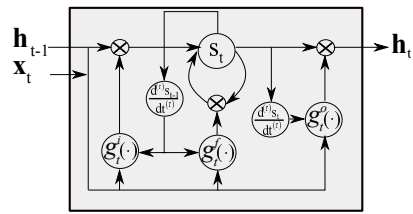}
\caption{Architecture of the differential recurrent neural network (dRNN) at time $t$. The input gate $i$ and the forget gate $f$ are controlled by the DoS at times $t-1$ and $t$, respectively, \cite{veeriah2015differential}.}
\label{fig:drnn}
\end{figure}

While LSTMs have shown improved learning ability in understanding long term sequential dependencies, it has been argued that its gating mechanisms have no way of comprehensively discriminating between salient and non-salient information in a sequence \cite{veeriah2015differential}. Therefore, LSTMs fail to capture spatio-temporal dynamic patterns in tasks such as action recognition \cite{veeriah2015differential}, in which sequences can often contain many non-salient frames. Differential recurrent neural networks (dRNNs) refer to detecting and capturing of important spatio-temporal sequences to learn dynamics of actions in input \cite{veeriah2015differential}. A LSTM gate in dRNNs monitors alternations in information gain of important motions between successive frames. This change of information is detectable by computing the derivative of hidden states (DoS). A large DoS reveals sudden change of actions state, which means the spatio-temporal structure contains informative dynamics. In this situation, the gates in Figure~\ref{fig:drnn} allow flow of information to update the memory cell defined as
\begin{equation}
\textbf{s}_{t}=\textbf{g}_{t}^{f}\odot \mathbf{s}_{t-1}+\textbf{g}_{i}^{t}\odot \mathbf{s}_{t-1/2}
\end{equation}
where
\begin{equation}
\mathbf{s}_{t-1/2}=tanh(\textbf{W}_{hs}\textbf{h}_{t-1}+\textbf{W}_{xs}\textbf{X}_{t}+\textbf{b}_{s}).
\end{equation}

\begin{table*}[]
	\centering
	\footnotesize
	\caption{A comparison between major long-short term memory (LSTM) architectures.}
	\begin{tabular}{|c|l|l|l|c|}
		\hline
		Method & \multicolumn{1}{c|}{Advantages}  & \multicolumn{1}{c|}{\begin{tabular}[c]{@{}l@{}}  Disadvantages \end{tabular}}  \\ \hline
		
		 LSTM   &  \begin{tabular}[c]{@{}l@{}} - models long-term dependencies better than a simple\\ RNN\\ - more robust to vanishing gradients than a simple \\RNN \end{tabular}  &  \begin{tabular}[c]{@{}l@{}}  - higher memory requirement and computational complexity \\than a simple RNN due to multiple memory cells \end{tabular}  \\ \hline
	
		S-LSTM   &  \begin{tabular}[c]{@{}l@{}} - models complicated inputs better than LSTM  \end{tabular}  &  \begin{tabular}[c]{@{}l@{}}  -  higher computational complexity in comparison with LSTM \end{tabular}  \\ \hline
		
		Stacked LSTM   &  \begin{tabular}[c]{@{}l@{}} -  models long-term sequential dependencies due to \\deeper architecture \end{tabular}  &  \begin{tabular}[c]{@{}l@{}}  - higher memory requirement and computational complexity \\than LSTM due to a stack of LSTM cells \end{tabular}  \\ \hline
		
		Bidirectional LSTM   &  \begin{tabular}[c]{@{}l@{}} - captures both future and past context of the input \\sequence better than LSTM and S-LSTM \end{tabular}  &  \begin{tabular}[c]{@{}l@{}}  - increases computational complexity in comparison with \\ LSTM due to the forward and backward learning  \end{tabular}  \\ \hline
		
		Multidimensional LSTM   &  \begin{tabular}[c]{@{}l@{}} - models multidimensional sequences \end{tabular}  &  \begin{tabular}[c]{@{}l@{}}  - higher memory requirement and computational complexity than\\ LSTM due to multiple hidden state vectors\\ - instability of the network as grid size and depth grows \end{tabular}  \\ \hline
		
		Grid LSTM   &  \begin{tabular}[c]{@{}l@{}} - models multidimensional sequences with \\ increased grid size  \end{tabular}  &  \begin{tabular}[c]{@{}l@{}} - higher memory requirement and computational complexity \\ than LSTM due to multiple recurrent connections \end{tabular}  \\ \hline
		
		Differential RNN   &  \begin{tabular}[c]{@{}l@{}} - discrimination between salient and non-salient\\ information in a sequence \\ -  better captures spatiotemporal patterns\end{tabular}  &  \begin{tabular}[c]{@{}l@{}}- increases computational complexity in comparison with LSTM\\ due to the differential operators \end{tabular}  \\ \hline

		Local-Global LSTM   &  \begin{tabular}[c]{@{}l@{}} - improves exploitation of local and global contextual\\ information in a sequence\end{tabular}  &  \begin{tabular}[c]{@{}l@{}} - increases computational complexity in comparison with LSTM \\due to more number of parameters for local and global \\representations \end{tabular}  \\ \hline		
		
		Matching LSTM   &  \begin{tabular}[c]{@{}l@{}} -  optimizes LSTM for natural language inference tasks\end{tabular}  &  \begin{tabular}[c]{@{}l@{}}  - increases computational complexity due to word-by-word\\ matching of hypothesis and premise \end{tabular}  \\ \hline
		
		Frequency-Time LSTM   &  \begin{tabular}[c]{@{}l@{}} - models both time and frequency \end{tabular}  &  \begin{tabular}[c]{@{}l@{}}  - more  computational complexity than LSTM due to more \\number of parameters to model time and frequency\end{tabular}  \\ \hline					
	\end{tabular}
	\label{T:lstms}
\end{table*}

The DoS $d\textbf{s}_{t}/d_{t}$ quantifies the change of information at each time $t$. Small DoS keeps the memory cell away from any influence by the input. More specifically, the cell controls the input gate as
\begin{equation}
\textbf{g}^{i}_{t} = \sigma(\sum_{r=0}^{R}\mathbf{W}_{dg^{i}}^{(r)} \frac{d^{(r)}\mathbf{s}_{t-1}}{dt^{(r)}} + \mathbf{W}_{hg^{i}} \mathbf{h}_{t-1} + \mathbf{W}_xg^{i} \mathbf{x}_{t} + \mathbf{b}_{g^{i}}),
\end{equation}
the forget gate unit as
\begin{equation}
\textbf{g}^{f}_{t} = \sigma(\sum_{r=0}^{R}\mathbf{W}_{dg^{f}}^{(r)} \frac{d^{(r)}\mathbf{s}_{t-1}}{dt^{(r)}} + \mathbf{W}_{hg^{f}} \mathbf{h}_{t-1} + \mathbf{W}_{xg^{f}} \mathbf{x}_{t} + \mathbf{b}_{g^{f}}),
\end{equation}
and the output gate unit as
\begin{equation}
\textbf{g}^{o}_{t} = \sigma(\sum_{r=0}^{R}\mathbf{W}_{dg^{o}}^{(r)} \frac{d^{(r)}\mathbf{s}_{t}}{dt^{(r)}} + \mathbf{W}_{hg^{o}} \mathbf{h}_{t-1} + \mathbf{W}_{xg^{o}} \mathbf{x}_{t} + \mathbf{b}_{g^{o}}),
\end{equation}
where the DoS has an upper order limit of $R$. BPTT can train dRNNs. The 1-order and 2-order dRNN have better performance in training comparing with the simple LSTM; however, it has additional computational complexity.

\subsubsection{Other LSTM Models}
The local-global LSTM (LG-LSTM) architecture is initially proposed for semantic object parsing \cite{Liang2015}. The objective is to improve exploitation of complex local (neighbourhood of a pixel) and global (whole image) contextual information on each position of an image. The current version of LG-LSTM has appended a stack of LSTM layers to intermediate convolutional layers. This technique directly enhances visual features and allows an end-to-end learning of network parameters \cite{Liang2015}. Performance comparison of LG-LSTM with a variety of CNN models show high accuracy performance~\cite{Liang2015}. It is expected that this model can achieve more success by replacing all convolutional layers with LG-LSTM layers.

The matching LSTM (mLSTM) is initially proposed for natural language inference. The matching mechanism stores (remembers) the critical results for the final prediction and forgets the less important matchings \cite{wang2015}. The last hidden state of the mLSTM is useful to predict the relationship between the premise and the hypothesis. The difference with other methods is that instead of a whole sentence embedding of the premise and the hypothesis, the mLSTM performs a word-by-word matching of the hypothesis with the premise \cite{wang2015}.

The recurrence in both time and frequency for RNN, named F-T-LSTM, is proposed in \cite{li2016}. This model generates a summary of the spectral information by scanning the frequency bands using a frequency LSTM. Then, it feeds the output layers activations as inputs to a LSTM. The formulation of frequency LSTM is similar to the time LSTM \cite{li2016}. A convolutional LSTM (ConvLSTM) model with convolutional structures in both the input-to-state and state-to-state transitions for precipitation now-casting is proposed in \cite{shi2015}. This model uses a stack of multiple ConvLSTM layers to construct an end-to-end trainable model \cite{shi2015}. A comparison between major LSTM models is provided in Table~\ref{T:lstms}.

\begin{figure}
\footnotesize
\centering
\includegraphics[width=0.5\linewidth]{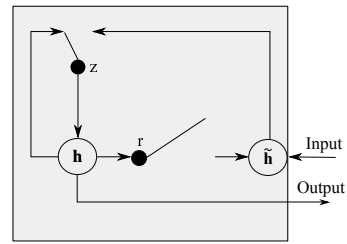}
\caption{A gated recurrent unit (GRU). The update gate $z$ decides if the hidden state is to be updated with
a new hidden state $\tilde h$. The reset gate $r$ controls if the previous hidden state needs to be ignored.}
\label{fig:GRU}
\end{figure}

\subsection{Gated Recurrent Unit}
While LSTMs have shown to be a viable option for avoiding vanishing or exploding gradients, they have a higher memory requirement given multiple memory cells in their architecture. Recurrent units adaptively capture dependencies of different time scales in gated recurrent units (GRUs)~\cite{chung2014empirical}. Similar to the LSTM unit, the GRU has gating units that modulate the flow of information inside the unit, however, without having separate memory cells. In contrast to LSTM, the GRU exposes the whole state at each timestep \cite{cho2014properties} and computes a linear sum between the existing state and the newly computed state. The block diagram of a GRU is presented in Figure~\ref{fig:GRU}. The activation in a GRU is linearly modeled as
\begin{equation}
\mathbf{h}_t=(1-z_{t})\mathbf{h}_{t-1}+z_{t}\mathbf{\tilde h}_{t},
\end{equation}
where the update gate $z_{t}$ controls update value of the activation, defined as
\begin{equation}
z_{t}=\sigma(\mathbf{W}_{z}\mathbf{x}_{t}+\textbf{U}_{z}\mathbf{h}_{t-1}),
\end{equation}
where $\mathbf{W}$ and $\textbf{U}$ are weight matrices to be learned. The candidate activation is
\begin{equation}
\mathbf{\tilde h}_{t}=tanh(\mathbf{W}_{h}\mathbf{x}_{t}+\textbf{U}_{h}(\mathbf{r}_{t}\odot \mathbf{h}_{t-1})),
\end{equation}
where $\mathbf{r}_{t}$ is a set of rest gates defined as
\begin{equation}
\textbf{r}_{t}=\sigma(\mathbf{W}_{r}\mathbf{x}_{t}+\textbf{U}_{r}\mathbf{h}_{t-1})
\end{equation}
which allows the unit to forget the previous state by reading the first symbol of an input sequence. Several similarities and differences between GRU networks and LSTM networks are outlined in \cite{chung2014empirical}. The study found that both models performed better than the other only in certain tasks, which suggests there cannot be a suggestion as to which model is better. 
%

\subsection{Memory Networks}
Conventional RNNs have small memory size to store features from past inputs \cite{weston2015}, \cite{weston2015a}. Memory neural networks (MemNN) utilize successful learning methods for inference with a readable and writable memory component. A MemNN is an array of objects and consists of input,
response, generalization, and output feature map components \cite{weston2015}, \cite{kumar2015}. It converts the input to an internal feature representation and then updates the memories based on the new input. Then, it uses the input and the updated memory to compute output features and decode them to produce an output \cite{weston2015}. This networks is not easy to train using BPTT and requires supervision at each layer \cite{Sukhbaatar2015}. A less supervision oriented version of MemNN is end-to-end MemNN, which can be trained end-to-end from input-output pairs \cite{Sukhbaatar2015}. It generates an output after a number of timesteps and the intermediary steps use memory input/output operations to update the internal state~\cite{Sukhbaatar2015}. 

Recurrent memory networks (RMN) take advantage of the LSTM as well as the MemNN~\cite{tran2016}. The memory block in RMN takes the hidden state of the LSTM and compares it to the most recent inputs using an attention mechanism.
The RMN algorithm analyses the attention weights of trained model and extracts knowledge from the retained information in the LSTM over time~\cite{tran2016}.
This model is developed for language modeling and is tested on three large datasets. The results show performance of the algorithm versus LSTM model, however, this model inherits the complexity of LSTM and RMN and needs further development.

Episodic memory is inspired from semantic and episodic memories, which are necessary for complex reasoning in the brain \cite{kumar2015}. Episodic memory is named as the memory of the dynamic memory network framework, which remembers autobiographical details \cite{kumar2015}. This memory refers to generated representation of stored experiential facts. The facts are retrieved from the inputs conditioned on the question. This results in a final representation by reasoning on the facts. The module performs several passes over the facts, while focusing on different facts. The output of each pass is called an episode, which is summarized into the memory \cite{kumar2015}. A relevant work to MemNN is the dynamic memory networks (DMN). An added memory component to the MemNN can boost its performance in learning long-term dependencies~\cite{weston2015}. This approach has shown performance for natural language question answering application~\cite{kumar2015}. The generalization and output feature map parts of the MemNN have some similar functionalities with the episodic memory in DMS. The MemNN processes sentences independently \cite{kumar2015}, while the DMS processes sentences via a sequence model \cite{kumar2015}. The performance results on the Facebook bAbI dataset show the DMN passes 18 tasks with accuracy of more than $95\%$ while the MemNN passes 16 tasks with lower accuracy \cite{kumar2015}. Several steps of Episodic memory are discussed in~\cite{kumar2015}.

\begin{figure}[!t]
	\footnotesize

	\centering
	\includegraphics[width=0.3\linewidth]{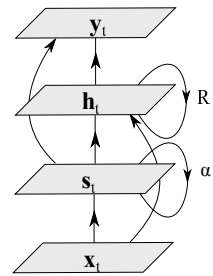}

	\caption{Recurrent neural network with context features (longer memory).}
	\label{fig:longer_mem}
\end{figure}

\begin{table*}[]
	\centering
	\footnotesize
	\caption{A comparison between major recurrent neural network (RNN) architectures.}
	\begin{tabular}{|c|l|l|l|c|}
		\hline
		Method & \multicolumn{1}{c|}{Advantages}  & \multicolumn{1}{c|}{\begin{tabular}[c]{@{}l@{}}  Disadvantages \end{tabular}}  \\ \hline
		
		Deep RNN &  \begin{tabular}[c]{@{}l@{}}  - disentangles variations of input sequence\\  - network can adapt to quick changing input nodes\\
- develops more compact hidden state\end{tabular}  &  \begin{tabular}[c]{@{}l@{}}  - increases computational complexity due to more number\\ of parameters comparing to a RNN\\ - deeper networks are more susceptible to vanishing of gradients \end{tabular}  \\ \hline
		
		\begin{tabular}[c]{@{}c@{}} Bidirectional \\RNN \end{tabular}  &  \begin{tabular}[c]{@{}l@{}}  - predicts both in the positive and negative time directions\\ simultaneously  \end{tabular}  &  \begin{tabular}[c]{@{}l@{}}  - must know both start and end of sequence\\ - increases computational complexity due to more number\\ of parameters comparing to a RNN\\\end{tabular}  \\ \hline
	
		\begin{tabular}[c]{@{}c@{}} Recurrent \\Convolutional \\Neural Network  \end{tabular}   &  \begin{tabular}[c]{@{}l@{}} - models long range spatial dependencies\\ -  embeds global spatial context into compact local \\representation \\ - activation evolves over time  \end{tabular}  &  \begin{tabular}[c]{@{}l@{}}  - increases computational complexity comparing to\\ a RNN\\ \end{tabular}  \\ \hline		
		
		\begin{tabular}[c]{@{}c@{}}Multi-Dimensional \\RNN  \end{tabular} &  \begin{tabular}[c]{@{}l@{}} - models high dimensional sequences\\ - more robust to warping than a RNN \end{tabular}  &  \begin{tabular}[c]{@{}l@{}}  - increases computational complexity comparing with\\ a RNN\\ - significantly increases memory requirements for training \\and testing due to multiple recurrent connections \end{tabular}  \\ \hline	
				
		\begin{tabular}[c]{@{}c@{}}Long-short \\term memory \\(LSTM)   \end{tabular} &  \begin{tabular}[c]{@{}l@{}} - capable of modeling long-term sequential dependencies\\ - more robust to vanishing gradients than a RNN \end{tabular}  &  \begin{tabular}[c]{@{}l@{}}  - increases computational complexity comparing with\\ a RNN\\ - higher memory requirement than RNN due to multiple\\ memory cells \end{tabular}  \\ \hline
		
		\begin{tabular}[c]{@{}c@{}}Gated \\Recurrent Unit   \end{tabular}  &  \begin{tabular}[c]{@{}l@{}} - capable of modeling long-term sequential dependencies\\ - more robust to vanishing gradients\\ - less memory requirements than LSTM  \end{tabular}  &  \begin{tabular}[c]{@{}l@{}}  - higher computational complexity and memory requirement\\ than a RNN due to multiple hidden state vectors 
		\end{tabular}  \\ \hline
		
		\begin{tabular}[c]{@{}c@{}}Recurrent Memory \\Networks  \end{tabular}   &  \begin{tabular}[c]{@{}l@{}} - capable of storing larger memory than a RNN\end{tabular}  &  \begin{tabular}[c]{@{}l@{}}  - increases memory requirements than a RNN\end{tabular}  \\ \hline
		
		\begin{tabular}[c]{@{}c@{}}Structurally\\ Constrained RNN  \end{tabular}   &  \begin{tabular}[c]{@{}l@{}} - stores larger memory than a RNN \\ - more robust to vanishing gradients than simple RNN \end{tabular}  &  \begin{tabular}[c]{@{}l@{}}  - not efficient in training \end{tabular}  \\ \hline	
		
		\begin{tabular}[c]{@{}c@{}}Unitary RNN   \end{tabular}  &  \begin{tabular}[c]{@{}l@{}} - models long-term sequential dependencies\\ - robustness to vanishing gradients\\ - less computational and memory requirements \\than gated RNN architectures \end{tabular}  &  \begin{tabular}[c]{@{}l@{}}  - requires more research and comparative study \end{tabular}  \\ \hline
		
		\begin{tabular}[c]{@{}l@{}}Gated Orthogonal \\Recurrent Unit\end{tabular} & \begin{tabular}[c]{@{}l@{}} - models long-term sequential dependencies\\ - robustness to vanishing gradients \end{tabular}  &  \begin{tabular}[c]{@{}l@{}}  - requires more research and comparative study \end{tabular}  \\ \hline
		
		\begin{tabular}[c]{@{}c@{}}Hierarchical\\ Subsampling\\ RNN \end{tabular}    &  \begin{tabular}[c]{@{}l@{}} - more robustness to vanishing gradients than a RNN \end{tabular}  &  \begin{tabular}[c]{@{}l@{}}  - sensitive to sequential distortions \\ - requires tuning window size \end{tabular}  \\ \hline	
	\end{tabular}
	\label{T:architectures}
\end{table*}

\subsection{Structurally Constrained Recurrent Neural Network}
Another model which aims to deal with the vanishing gradient problem is the structurally constrained recurrent neural network (SCRN). This network is based on the observation that the hidden states change rapidly during training, as presented in Figure~\ref{fig:longer_mem},~\cite{mikolov2014learning}. In this approach, the SCRN structure is extended by adding a specific recurrent matrix equal to identity longer term dependencies. The fully connected recurrent matrix (called hidden layer) produces a set of quickly changing hidden units, while the diagonal matrix (called context layer) supports slow change of the state of the context units \cite{mikolov2014learning}. In this way, state of the hidden layer stays static and changes are fed from external inputs. Although this model can prevent gradients of the recurrent matrix vanishing, it is not efficient in training \cite{mikolov2014learning}. In this model, for a dictionary of size $d$, $\textbf{s}_{t}$ is the state of the context units defined as
\begin{equation}
\textbf{s}_{t}=(1-\alpha)\textbf{B}\textbf{x}_{t}+\alpha \textbf{s}_{t-1},
\end{equation}
where $\alpha$ is the context layer weight, normally set to 0.95, $\textbf{B}_{d\times s}$ is the context embedding matrix, and $\textbf{x}_{t}$ is the input. The hidden layer is defined as
\begin{equation}
\textbf{h}_{t}=\sigma (\textbf{P}\textbf{s}_{t}+\textbf{A}\textbf{x}_{t}+\textbf{R}\textbf{h}_{t-1}),
\end{equation}
where $\textbf{A}_{d\times m}$ is the token embedding matrix, $\textbf{P}_{p\times m}$ is the connection matrix between hidden and context layers, $\textbf{R}_{m\times m}$ is the hidden layer $\textbf{h}_{t-1}$ weights matrix, and $\sigma(\cdot)$ is the ``sigmoid" activation function. Finally, the output $\textbf{y}_{t}$ is defined as
\begin{equation}
\textbf{y}_{t}=f(\textbf{U}\textbf{h}_{t}+\textbf{V}\textbf{s}_{t}),
\end{equation}
where $f$ is the ``softmax" activation function, and $\textbf{U}$ and $\textbf{V}$ are the output weight matrices of hidden and context layers, respectively.

Analysis using adaptive context features, where the weights of the context layer are learned for each unit to capture context from different time delays, show that learning of the self-recurrent weights does not seem to be important, as long as one uses also the standard hidden layer in the model. This is while fixing the weights of the context layer to be constant, forces the hidden units to capture information on the same time scale.
The SCRN model is evaluated on the Penn Treebank dataset. The presented results in \cite{mikolov2014learning} show that the SCRN method has bigger gains compared to the proposed model in~\cite{bengio2013advances}. Also, the learning longer memory model claims that it has similar performance, but with less complexity, comparing to the LSTM model~\cite{mikolov2014learning}.

While adding the simple constraint to the matrix results in lower computation compared to its gated counterparts, the model is not efficient in training. The analysis of using adaptive context features, where the weights of the context layer are learned for each unit to capture context from different time delays, shows that learning of the self-recurrent weights does not seem to be important, as long as one uses also the standard hidden layer in the model \cite{mikolov2014learning}. Thus, fixing the weights of the context layer to be constant forces the hidden units to capture information on the same time scale.

\subsection{Unitary Recurrent Neural Networks}
A simple approach to alleviating the vanishing and exploding gradients problem is to simply use unitary matrices in a RNN. The problem of vanishing or exploding gradients can be attributed to the eigenvalues of the hidden to hidden weight matrix, deviating from one \cite{arjovsky2016unitary}. Thus, to prevent these eigenvalues from deviating, unitary matrices can be used to replace the general matrices in the network. 

Unitary matrices are orthogonal matrices in the complex domain \cite{arjovsky2016unitary}. They have absolute eigenvalues of exactly one, which preserves the norm of vector flows and the gradients to propagate through longer timesteps. This leads to  preventing vanishing or exploding gradient problems from arising \cite{jing2017gated}. However, it has been argued that the ability to back propagate gradients without any vanishing could lead to the output being equally dependent on all inputs regardless of the time differences \cite{jing2017gated}. This also results in the network to waste memory due to storing redundant information.

Unitary RNNs have significant advantages over previous architectures, which have attempted to solve the vanishing gradient problem. A unitary RNN architecture keeps the internal workings of a vanilla RNN without adding any additional memory requirements. Additionally, by maintaining the same architecture, Unitary RNNs do not noticeably increase the computational cost.

\subsection{Gated Orthogonal Recurrent Unit}
Thus far, implementations of RNNs have taken two separate approaches to tackle the issues of exploding and vanishing gradients. The first is implementation of additional gates to improve memory of the system, such as the LSTM and GRU architectures. The second is implementation of unitary matrices to maintain absolute eigenvalues of one. 

The gated orthogonal recurrent unit replaces the hidden state loop matrix with an orthogonal matrix and introduces an augmentation of the ReLU activation function, which allows it to handle complex-value inputs \cite{jing2017gated}. This unit is capable of capturing long term dependencies of the data using unitary matrices, while also leverages forgetting mechanisms present in the GRU structure \cite{jing2017gated}.

\subsection{Hierarchical Subsampling Recurrent Neural Networks}
It has been shown that RNNs particularly struggle with learning long sequences. While previous architectures have aimed to change the mechanics of the network to better learn long term dependencies, a simpler solution exists, shortening the sequences using methods such as subsampling. Hierarchical subsampling recurrent neural networks (HSRNNs) aim to better learn large sequences by performing subsampling at each level using a fixed window size \cite{graves2012supervised}. Training this network follows the same process as training a regular RNN, with a few modifications based on the window sizes at each level.

HSRNNs can be extended to multidimensional networks by simply replacing the subsampling windows with multidimensional windows~\cite{graves2012supervised}. In multidirectional HSRNNs, each level consists of two recurrent layers scanning in two separate directions with a feedforward layer in between. However, in reducing the size of the sequences, the HSRNN becomes less robust to sequential distortions. This requires a lot of tuning of the network than other RNN models, since the optimal window size varies depending on the task~\cite{graves2012supervised}. HSRNNs have shown to be a viable option as a model for learning long sequences due to their lower computational costs when compared to their counterparts. RNNs, regardless of their internal architecture, are activated at each time step of the sequence. This can cause extremely high computational costs for the network to learn information in long sequences \cite{graves2012supervised}. Additionally, information can be widely dispersed in a long sequence, making inter-dependencies harder to find. A comparison between major RNN architectures is provided in Table~\ref{T:architectures}.

 \section{Regularizing Recurrent Neural Networks}
\label{sec:regularization}
 
\begin{figure}[!t]
 	\footnotesize
 	\centering
 	\includegraphics[width=0.7\linewidth]{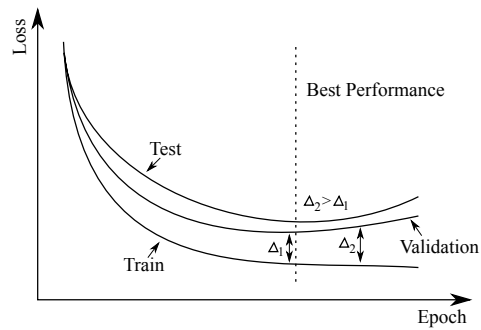}
 	
 	\caption{Overfitting in training neural networks. To avoid overfitting, it is possible to early-stop the training at the ``Best Performance" epoch, where the training loss is decreasing but the validation loss starts increasing.  }
 	\label{Fig:overfitting}     
 \end{figure}

Regularization refers to controlling the capacity of the neural network by adding or removing information to prevent overfitting. For better training of a RNN, a portion of available data is considered as validation dataset. The validation set is used to watch the training procedure and prevent the network from underfitting and overfitting \cite{bishop2006pattern}. Overfitting refers to the gap between the training loss and the validation loss (including the test loss), which increases after a number of training epochs as the training loss decreases, as demonstrated in Figure~\ref{Fig:overfitting}. Successful training of RNNs requires good regularization~\cite{srivastava2013improving}. This section aims to introduce common regularization methods in training RNNs.

 \subsection{$L_{1}$ and $L_{2}$}

The $L_{1}$ and $L_{2}$ regularization methods add a regularization term to the loss function to penalize certain parameter configuration and prevent the coefficients from fitting so perfectly as to overfit. The loss function in Eq. (\ref{eq:loss_function}) with added regularization term is
\begin{equation}
 \mathcal{L}(\textbf{y}, \textbf{z})= \mathcal{L}(\textbf{y}, \textbf{z}) + \eta \norm{\theta}^{p}_{p},
\label{}
\end{equation}
where $\theta$ is the set of network parameters (weights), $\eta$ controls the relative importance of the regularization parameter, and
\begin{equation}
\norm{\theta}_{p}=(\sum_{j=0}^{|\theta|}|\theta_{j}|^{p})^{1/p}.
\end{equation}
If $p=1$ the regularizer is $L_{1}$ and if $p=2$ the regularizer is $L_{2}$.
The $L_{1}$ is the sum of the weights and $L_{2}$ is the sum of the square of the weights.

 
 \subsection{Dropout}
In general, the dropout randomly omits a fraction of the connections between two layers of the network during training. For example, for the hidden layer outputs in Eq.~(\ref{eq:SRNN_hidden_state}) we have
 \begin{equation}
 \mathbf{h}_{t} = \mathbf{k} \odot  \mathbf{h}_{t},
 \end{equation} 
 where $\mathbf{k}$ is a binary vector mask and $\odot$ is the element-wise product \cite{pham2014dropout}. The mask can also follow a statistical pattern in applying the withdrawal. During testing, all units are retained and their activations may be weighted.
 
 
A dropout specifically tailored to RNNs is introduced in \cite{moon2015rnndrop}, called RNNDrop. This method generates a single dropout mask at the beginning of each training sequence and adjusts it over the duration of the sequence. This allows the network connections to remain constant through time. Other implementations of dropout for RNNs suggest simply dropping the previous hidden state of the network. A similar model to the RNNDrop is introduced in \cite{semeniuta2016recurrent}, where instead of dropout, it masks data samples at every input sequence per step. This small adjustment has competitive performance to the RNNDrop.
 

\begin{figure}[!t]
	\footnotesize
	
	\centering 
	\includegraphics[width=0.9\linewidth]{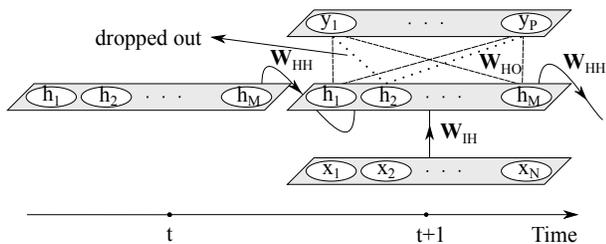}
	
	\caption{Dropout applied to feed-forward connections in a RNN. The recurrent connections are shown as full connection with a solid line. The connection between hidden units and output units are shown in dashed lines. The dropped-out connection between the hidden units and output units are shown by dotted lines.}
	\label{fig:dropout}    
	\vspace{-5mm} 
\end{figure}

\subsection{Activation Stabilization}
Another recently proposed method of regularization involves stabilizing the activations of the RNNs \cite{krueger2015regularizing}. The norm-stabilizer is an additional cost term to the loss function defined as
 \begin{equation}
\mathcal{L}(\textbf{y}, \textbf{z})= \mathcal{L}(\textbf{y}, \textbf{z}) + \beta \frac{1}{T} \sum^{T}_{t=1}(||\textbf{h}_t||_{2} - ||\textbf{h}_{t-1}||_2)^2 
\end{equation} 
where $\textbf{h}_t$ and  $\textbf{h}_{t-1}$ are the vectors of the hidden activations at time $t$ and $t-1$, respectively, and $\beta$ controls the relative importance of the regularization. This additional term stabilizes the norms of the hidden vectors when generalizing long-term sequences. 

Other implementations have been made to stabilize the hidden-to-hidden transition matrix such as the use of orthogonal matrices, however, inputs and nonlinearities can still affect the stability of the activation methods. Experiments  on language modelling and phoneme recognition show state of the art performance of this approach \cite{krueger2015regularizing}. 

\subsection{Hidden Activation Preservation}
The zoneout method is a very special case of dropout. It forces some units to keep their activation from the previous timestep (i.e., $\mathbf{h}_{t}= \mathbf{h}_{t-1}$) \cite{krueger2016zoneout}. This approach injects stochasticity (by adding noise) into the network, which makes the network more robust to changes in the hidden state and help the network to avoid overfitting. Zoneout uses a Bernoulli mask $\mathbf{k}$ to modify dynamics of $\mathbf{h}_{t}$ as 
 \begin{equation}
 \mathbf{h}_{t} = \mathbf{k} \odot  \mathbf{h}_{t} + (1- \mathbf{k})  \odot  1
 \end{equation} 
which improves the flow of information in the
network~\cite{krueger2016zoneout}. Zoneout has slightly better performance than dropout. However, it can also work together with dropout and other regularization methods~\cite{krueger2016zoneout}.

\section{Recurrent Neural Networks for Signal Processing}
\label{sec:RNNsigapps}
RNNs have various applications in different fields and a large number of research articles are published in that regard. In this section, we review different applications of RNNs in signal processing, particularly text, audio and speech, image, and video processing.

\subsection{Text}
RNNs are developed for various application in natural language processing and language modeling. 
RNNs can outperform \textit{n}-gram models and are widely used as language modelers~\cite{mikolov2011empirical}. However, RNNs are computationally more expensive and challenging to train. A method based on factorization of output layer is proposed in~\cite{mikolov2011extensions}, which can speed-up the training of a RNN for language modeling up to 100 times. 
In this approach, words are assigned to specific categories based on their unigram frequency and only words belonging to the predicted class are evaluated in the output layer~\cite{mikolov2011empirical}.
HF optimization is used in \cite{sutskever2011generating} to train RNNs for character-level language modeling. This model uses gated connections to allow the current input character to determine the transition matrix from one hidden state vector to the next~\cite{sutskever2011generating}. 
LSTMs have improved RNN models for language modeling due to their  due to their ability to learn long-term dependencies in a sequence better than a simple hidden state~\cite{sundermeyer2012lstm}.
LSTMs are also used in~\cite{graves2013generating} to generate complex text and online handwriting sequences with long-range
structure, simply by predicting one data point at a time. 
RNNs are also used to capture poetic style in works of literature and generate lyrics, for example Rap lyric generation~\cite{zhang2014chinese, potash2015ghostwriter, ghazvininejad2016generating}.
A variety of document classification tasks is proposed in the literature using RNNs. In~\cite{tang2015document}, a GRU is adapted to perform document level sentiment analysis. In~\cite{lai2015recurrent}, RCNNs are used for text classification on several datasets. In such approaches, generally the words are mapped to a feature vector and the sequence of feature vectors are passed as input to the RNN model. 
The same sequence of feature vectors can also be represented as a feature matrix (i.e., an image) to be fed as input to a CNN.
CNNs are used in~\cite{salehinejad2017interpretation} to classify radiology reports. The
proposed model is particularly developed for chest pathology and mammogram reports.
However, RNNs have not yet been examined for medical reports interpretation and can
potentially result in very high classification performance.

\subsection{Speech and Audio}
Speech and audio signals continuously vary over time. The inherent sequential and time varying nature of audio signals make RNNs the ideal model to learn features in this field. 


Until recently, RNNs had limited contribution in labelling unsegmented speech data, primarily because this task requires pre-segmented data and post-processing to produce outputs~\cite{graves2006connectionist}. Early models in speech recognition, such as time-delay neural networks, often try to make use of the sequential nature of the data by feeding an ANN a set of frames~\cite{waibel1989phoneme}. Given that both past and future sequential information can be of use in speech recognition predictions, the concept of BRNNs were introduced for speech recognition~\cite{schuster1996bi}. Later, RNNs were combined with hidden Markov models (HMM) in which the HMM acted as an acoustic model while the RNN acted as the language model~\cite{bourlard2012connectionist}. With the introduction of the connectionist temporal classification (CTC) function, RNNs are capable of leveraging sequence learning on unsegmented speech data \cite{graves2006connectionist}. Since then, the popularity of RNNs in speech recognition has exploded. Developments in speech recognition then used the CTC function alongside newer recurrent network architectures, which were more robust to vanishing gradients to improve performance and perform recognition on larger vocabularies~\cite{graves2014towards,sak2014long,bahdanau2016end}.  Iterations of the CTC model, such as the sequence transducer and neural transducer~\cite{graves2013generating,jaitly2015neural} have incorporated a second RNN to act as a language model to tackle tasks such as online speech recognition. These augmentations allows the model to make predictions based on not only the linguistic features, but also on the previous transcriptions made. 


Speech emotion recognition is very similar to speech recognition, such that a segment of speech must be classified as an emotion. Thus the development of speech emotion recognition followed the same path as that of speech recognition. HMMs were initially used for their wide presence in speech applications \cite{el2011survey}. Later, Gaussian mixture models (GMMs) were adapted to the task for their lower training requirements and efficient multi-modal distribution modeling \cite{el2011survey}. However, these models often require hand crafted and feature engineered input data. Some examples are mel-frequency cepstral coefficients (MFCCs), perceptual linear prediction (PLP) coefficients, and supra-segmental features \cite{trigeorgis2016adieu}. With the introduction of RNNs, the trend of input data began to shift from such feature engineering to simply feeding the raw signal as the input, since the networks themselves were able to learn these features on their own. Several RNN models have been introduced since then to perform speech emotion recognition. In \cite{wollmer2008abandoning}, an LSTM network is shown to have better performance than support vector machines (SVMs) and conditional random fields (CRFs). This improved performance is attributed to the network's ability to capture emotions by better modeling long-range dependencies. In \cite{lee2015high}, a deep BLSTM is introduced for speech emotion recognition. Deep BLSTMs are able to capture more information by taking in larger number of frames while a feed-forward DNN simply uses the frame with the highest energy in a sequence \cite{lee2015high}. However, comparisons to previous RNNs used for speech emotion recognition were not made. Given that this model used a different model than the LSTM model described prior, no comparisons could be found as to which architecture performs better. Recently, a deep convolutional LSTM is adapted in \cite{trigeorgis2016adieu}. This model gives state-of-the-art performance when tested on the RECOLA dataset, as the convolutional layers learns to remove background noise and outline important features in the speech, while the LSTM models the temporal structure of the speech sequence. 


Much like speech recognition, speech synthesis also requires long-term sequence learning. HMM-based models can often produce synthesized speech, which does not sound natural. This is due to the overly smooth trajectories produced by the model, as a result of statistical averaging during the training phase \cite{fan2015fast}. Recent advancements in ANNs have shown that deep MLP neural networks can synthesize speech. However, these models take each frame as an independent entity from its neighbours and fail to take into account the sequential nature of speech~\cite{fan2015fast}. RNNs were first used for speech synthesis to leverage these sequential dependencies \cite{karaali1998text,tuerk1993speech}, and were then replaced with LSTM models to better learn long term sequential dependencies \cite{zen2015unidirectional}. The BLSTM has been shown to perform very well in speech synthesis due to the ability to integrate the relationship with neighbouring frames in both future and past time steps \cite{fan2014tts,fernandez2014prosody}. CNNs have shown to perform better than state of the art LSTM models, in particular the WaveNet model \cite{oord2016wavenet}. WaveNet is a newly introduced CNN capable of generating speech, using dilated convolutions. Through the use of dilated causal convolutions, WaveNet can model long-range temporal dependencies by increasing it's receptive field of input. WaveNet has shown better performance than LSTMs and HMMs~\cite{oord2016wavenet}.

The modelling of polyphonic music presents another task with inherent contextual dependencies. In \cite{boulanger2012modeling}, a RNN combined with a restricted Boltzmann machine (RBM) is introduced, which is capable of modeling temporal information in a music track. This model has a sequence of conditional RBMs, which are fed as parameters to a RNN, so that can learn harmonic and rhythmic probability rules from polyphonic music of varying complexity~\cite{boulanger2012modeling}. It has been shown that RNN models struggle to keep track of distant events that indicate the temporal structure of music \cite{eck2002first}. LSTM models have since been adapted in music generation to better learn the long-term temporal structure of certain genres of music \cite{eck2002first,eck2002finding}.

\subsection{Image}
Learning the spatial dependencies is generally the main focus in machine vision. While CNNs have dominated most applications in computer vision and image processing, RNNs have also shown promising results such as image labeling, image modeling, and handwriting recognition. 

Scene labeling refers to the task of associating every pixel in an image to a class. This inherently involves the classification of a pixel to be associated with the class of its neighbour pixels. However, models such as CNNs have not been completely successful in using these underlying dependencies in their model. These dependencies have been shown to be leveraged in numerous implementations of RNNs. A set of images are represented as undirected cyclic graphs (UCGs) in~\cite{shuai2016dag}. To feed these images into a RNN, the UCGs are decomposed into several directed acyclic graphs (DAGs) meant to approximate the original images. Each DAG image involves a convolutional layer to produce discriminative feature mapping, a DAG-RNN to model the contextual dependencies between pixels, and a deconvolutional layer to up-sample the feature map to its original image size. This implementation has better performance than other state of the art models on popular datasets such as SiftFlow, CamVid, and Barcelona~\cite{shuai2016dag}. A similar implementation is shown in \cite{shuai2015quaddirectional}, where instead of decomposing the image into several DAGs, the image is first fed into a CNN to extract features for a local patch, which is then fed to a 2D-RNN. This 2D-RNN is similar to a simple RNN, except for its ability to store hidden states in two dimensions. The two hidden neurons flow in different directions towards the same neuron to create the hidden memory.
To encode the entire image, multiple starting points are chosen to create the multiple hidden states of the same pixel. This architecture is developed further by introducing 2D-LSTM units to better retain long-term information~\cite{byeon2015scene}.


 Image modeling is the task of assigning a probability distribution to an image. RNNs are naturally the best choice for image modeling tasks given its inherent ability to be used as a generative model. 
The deep recurrent attentive writer (DRAW) combines a novel spatial attention mechanism that mimics the foveation of the human eye, with a sequential variational auto-encoding framework that allows iterative construction of complex images~\cite{gregor2015draw}. While most image generative models aim to generate scenes all at once, this causes all pixels to be modelled on a single latent distribution. The DRAW model generates images through first generating sections of the scene independently of each other before going through iterations of refinement. The recent introduction of PixelRNN, involving LSTMs and BLSTMs, has shown improvements in modelling natural images with scalability~\cite{oord2016pixel}. The PixelRNN uses up to 12 2-dimensional LSTM layers, each of which has an input-to-state component and a recurrent state-to-state component. These component then determine the gates inside each LSTM. To compute these states, masked convolutions are used to gather states along one of the dimensions of the image. This model has better log-likelihood scores than other state of the art models evaluated on MNIST and CIFAR-10 datasets. While PixelRNN has shown to perform better than DRAW on the MNIST dataset, there has been no comparison between the two models as to why this might be. 


The task of handwriting recognition combines both image processing and sequence learning. This task can be divided into two types, online and offline recognition. RNNs perform well on this task, given the contextual dependencies in letter sequences \cite{graves2008unconstrained}. For the task of online handwriting recognition, the position of the pen-tip is recorded at intervals and these positions are mapped to the sequence of words~\cite{graves2008unconstrained}. In~\cite{graves2008unconstrained}, a BLSTM model is introduced for online handwriting recognition. Performance of this model is better than conventionally used HMM models due to its ability to make use of information in both past and future time steps. BLSTM perform well when combined with a probabilistic language model and trained with CTC. For offline handwriting recognition, only the image of the handwriting is available. To tackle this problem, an MDLSTM is used to convert the 2-dimensional inputs into a 1-dimensional sequence~\cite{graves2009offline}. The data is then passed through a hierarchy of MDLSTMs, which incrementally decrease the size of the data. While such tasks are often implemented using CNNs, it is argued that due to the absence of recurrent connections in such networks, CNNs cannot be used for cursive handwriting recognition without first being pre-segmented~\cite{graves2009offline}. The MDLSTM model proposed in~\cite{graves2009offline} offers a simple solution which does not need segmented input and can learn the long-term temporal dependencies. 

Recurrent generative networks are developed in~\cite{mardani2017recurrent} to automatically recover images from compressed linear
measurements. In this model, a novel proximal learning framework is developed, which adopts ResNets to model the proximals and a mixture of pixel-wise and perceptual costs are used for training. The deep convolutional generative adversarial networks are developed in~\cite{salehinejad2017generalization} to generate artificial chest radiographs for automated abnormality detection in chest radiographs. This model can be extended to medical image modalities which have spatial and temporal dependencies, such as head magnetic resonance imaging (MRI), using RCNNs. Since RNNs can model non-linear dynamical systems, recent RNN architectures can potentially enhance performance of these models. 

\subsection{Video}
A video is a sequence of images (i.e., frames) with temporal and spatial dependencies between frames and pixels in each frame, respectively. A video file has far more pixels in comparison to a single image, which results in a greater number of parameters and computational cost to process it. While different tasks have been performed on videos using RNNs, they are most prevalent in video description generation. This application involves components of both image processing and natural language processing. The method proposed in \cite{donahue2015long} combines a CNN for visual feature extraction with an LSTM model capable of decoding the features into a natural language string known as long-term recurrent convolutional
networks (LRCNs). However, this model was not an end-to-end solution and required supervised intermediate representations of the features generated by the CNN. This model is built upon in \cite{venugopalan2014translating}, which introduces a solution capable of being trained end-to-end. This model utilizes an LSTM model, which directly connects to a deep CNN. This model was further improved in \cite{yao2015video}, in which a 3-dimensional convolutional architecture was introduced for feature extraction. These features were then fed to an LSTM model based on a soft-attention mechanism to dynamically control the flow of information from multiple video frames. RNNs has fewer advances in video processing, comparing with the other types of signals, which introduces new opportunities for temporal spatial machine learning.  


%

\section{Conclusion and Potential directions}
\label{sec:conclusion}
In this paper, we systematically review major and recent advancements of RNNs in the literature and introduce the challenging problems in training RNNs. A RNN refers to a network of artificial neurons with recurrent connections among them. The recurrent connections learn the dependencies among input sequential or time-series data. The ability to learn sequential dependencies has allowed RNNs to gain popularity in applications such as speech recognition, speech synthesis, machine vision, and video description generation. 

One of the main challenges is training RNNs is learning long-term dependencies in data. It occurs generally due to the large number of parameters that need to be optimized during training in RNN over long periods of time. This paper discusses several architectures and training methods that have been developed to tackle the problems associated with training of RNNs. The followings are some major opportunities and challenges in developing RNNs: 
\begin{itemize}
	\item The introduction of BPTT algorithm has facilitated efficient training of RNNs. However, this approaches introduces gradient vanishing and explodin problems. Recent advances in RNNs have since aimed at tackling this issue. However, these challenges are still the main bottleneck of training RNNs.
	\item Gating mechanisms have been a breakthrough in allowing RNNs to learn long-term sequential dependencies. Architectures such as the LSTM and GRU have shown significantly high performance in a variety of applications. However, these architectures introduce higher complexity and computation than simple RNNs. Reducing the internal complexity of these architectures can help reduce training time for the network. 
	\item The unitary RNN has potentially solved the above issue by introducing a simple architecture capable of learning long-term dependencies. By replacing the internal weights with unitary matrices, the architecture keeps the same complexity of a simple RNN while providing stronger modeling ability. Further research into the use of unitary RNNs can help in validating its performance against its gated RNN counterparts. 
	\item Several regularization methods such as dropout, activation stabilization, and activation preservation have been adapted for RNNs to avoid overfitting. While these methods have shown to improve performance, there is no standard for regularizing RNNs. Further research into RNNs regularization can help introduce potentially better regularization methods. 
	\item RNNs have a great potential to learn features from 3-dimensional medical images, such as head MRI scans, lung computed tomography (CT), and abdominal MRI. In such modalities, the temporal dependency between images is very important, particularly for cancer detection and segmentation. 
\end{itemize}


\bibliographystyle{IEEEtran}
\bibliography{CTLIEEEtrans,mybibfile}

\end{document}